\documentclass[dvipsnames,format=sigconf,anonymous=false,review=false]{acmart}

\usepackage{amsmath,amsfonts}
\usepackage{algorithm}
\usepackage{array}
\usepackage[caption=false,font=normalsize,labelfont=sf,textfont=sf]{subfig}
\usepackage{textcomp}
\usepackage{stfloats}
\usepackage{url}
\usepackage{verbatim}
\usepackage{graphicx}
\usepackage{cite}
\usepackage{bbm}
\setcopyright{none}  % Remove copyright notice
\acmConference[Preprint]{}{accepted to Gecco '26, San José, Costa Rica, July 13 - 17}{DOI: 10.1145/3795095.3805154}
\renewcommand\footnotetextcopyrightpermission[1]{}

\usepackage{url, hyperref, cleveref}
\hypersetup{
    colorlinks=true,
    linkcolor=blue,
    filecolor=magenta,      
    urlcolor=blue,
    citecolor=blue,
}
\usepackage{booktabs}
\usepackage[dvipsnames]{xcolor}
\usepackage{float}
\usepackage{enumitem}
\usepackage[noend]{algpseudocode}
\usepackage{amsmath}
\usepackage{siunitx}
\usepackage{multirow}
\usepackage{comment}
\usepackage[table,xcdraw]{xcolor}
\usepackage{makecell}
\usepackage{nicematrix}
\usepackage{dsfont}
\usepackage{placeins}  % in preamble

\algrenewcommand{\algorithmiccomment}[1]{\hfill\colorbox{light-gray}{$\triangleright$ #1}}

\definecolor{light-gray}{gray}{0.92}
\definecolor{dark-gray}{gray}{0.85}
\definecolor{blue}{HTML}{2070b4}
\definecolor{red}{HTML}{ca171c}

\begin{document}
%%%%%%%%%%%%%%%%%%%%%%%%%%%%%%%%%%%%%%%%%%%%%%%%%%%%%%%%%%%%%%%%%%%%%%%%%%%%%%%%
%   Title
%%%%%%%%%%%%%%%%%%%%%%%%%%%%%%%%%%%%%%%%%%%%%%%%%%%%%%%%%%%%%%%%%%%%%%%%%%%%%%%%
\title{Tournament Informed Adversarial Quality Diversity} 

\author{Timothée Anne}
\orcid{0000-0002-4805-0213} 
\email{timl@itu.dk}
\affiliation{%
  \institution{IT University of Copenhagen}
  \city{Copenhagen}
  \country{Denmark}
}

\author{Noah Syrkis} 
\email{nobr@itu.dk}
\orcid{0000-0002-6553-1115} 
\affiliation{%
  \institution{IT University of Copenhagen}
  \city{Copenhagen}
  \country{Denmark}
}

\author{Meriem Elhosni}
\email{meriem.elhosni@ar.admin.ch}
\orcid{0009-0001-8552-6057}
\affiliation{%
  \institution{armasuisse Science+Technology}
  \city{Thun}
  \country{Switzerland}
}

\author{Florian Turati}
\email{florian.turati@ar.admin.ch}
\orcid{0009-0009-4729-7394} 
\affiliation{%
  \institution{armasuisse Science+Technology}
  \city{Thun}
  \country{Switzerland}
}

\author{Alexandre Manai}
\email{alexandre.manai@ar.admin.ch}
\orcid{0009-0001-9499-4776} 
\affiliation{%
  \institution{armasuisse Science+Technology}
  \city{Thun}
  \country{Switzerland}
}

\author{Franck Legendre}
\email{franck.legendre@ar.admin.ch}
\orcid{0009-0003-1169-4693}
\affiliation{%
  \institution{armasuisse Science+Technology}
  \city{Thun}
  \country{Switzerland}
}

\author{Alain Jaquier}
\email{alain.jaquier@ar.admin.ch}
\orcid{0009-0000-6507-4025}
\affiliation{%
  \institution{armasuisse Science+Technology}
  \city{Thun}
  \country{Switzerland}
}

\author{Sebastian Risi} 
\email{sebr@itu.dk}
\orcid{0000-0003-3607-8400} 
\affiliation{%
  \institution{IT University of Copenhagen}
  \city{Copenhagen}
  \country{Denmark}
}

\renewcommand{\shortauthors}{Anne et al.}

%%%%%%%%%%%%%%%%%%%%%%%%%%%%%%%%%%%%%%%%%%%%%%%%%%%%%
%   Abstract
%%%%%%%%%%%%%%%%%%%%%%%%%%%%%%%%%%%%%%%%%%%%%%%%%%%%%%
\begin{abstract}
Quality Diversity (QD) is a branch of evolutionary computation that seeks high-quality and behaviorally diverse solutions to a problem. While adversarial problems are common, classical QD cannot be easily applied to them, as both fitness and behavior depend on the opposing solutions. Recently, Generational Adversarial MAP-Elites (GAME) has been proposed to coevolve both sides of an adversarial problem by alternating the execution of a multi-task QD algorithm against previous elites, called tasks. The original algorithm selects new tasks based on a behavioral criterion, which may lead to undesired dynamics due to inter-side dependencies.  
In addition, comparing sets of solutions cannot be done directly using classical QD metrics due to inter-side dependencies. In this paper, we propose (1) 6 metrics of adversarial quality and diversity based on an inter-variants tournament to compare the sets of solutions, ensuring a fair comparison, and (2) propose two tournament-informed task selection methods to promote higher quality and diversity at each generation. We evaluate the variants across three adversarial problems: Pong, a Cat-and-mouse game, and a Pursuers-and-evaders game. We show that the tournament-informed task selection method leads to higher adversarial quality and diversity. We hope that this work will help further advance adversarial QD.
\end{abstract}  

%%%%%%%%%%%%%%%%%%%%%%%%%%%%%%%%%%%%%%%%%%%%%%%%%%%%%%
%   Keywords 
%%%%%%%%%%%%%%%%%%%%%%%%%%%%%%%%%%%%%%%%%%%%%%%%%%%%%%

\maketitle

%%%%%%%%%%%%%%%%%%%%%%%%%%%%%%%%%%%%%%%%%%%%%%%%%%%%%%%%%%%%%%%%%%%%%%%%%%%%%%%%
%%%   Introduction 
%%%%%%%%%%%%%%%%%%%%%%%%%%%%%%%%%%%%%%%%%%%%%%%%%%%%%%%%%%%%%%%%%%%%%%%%%%%%%%%%
\section{Introduction}

\begin{figure}
    \centering
    \includegraphics[width=\linewidth]{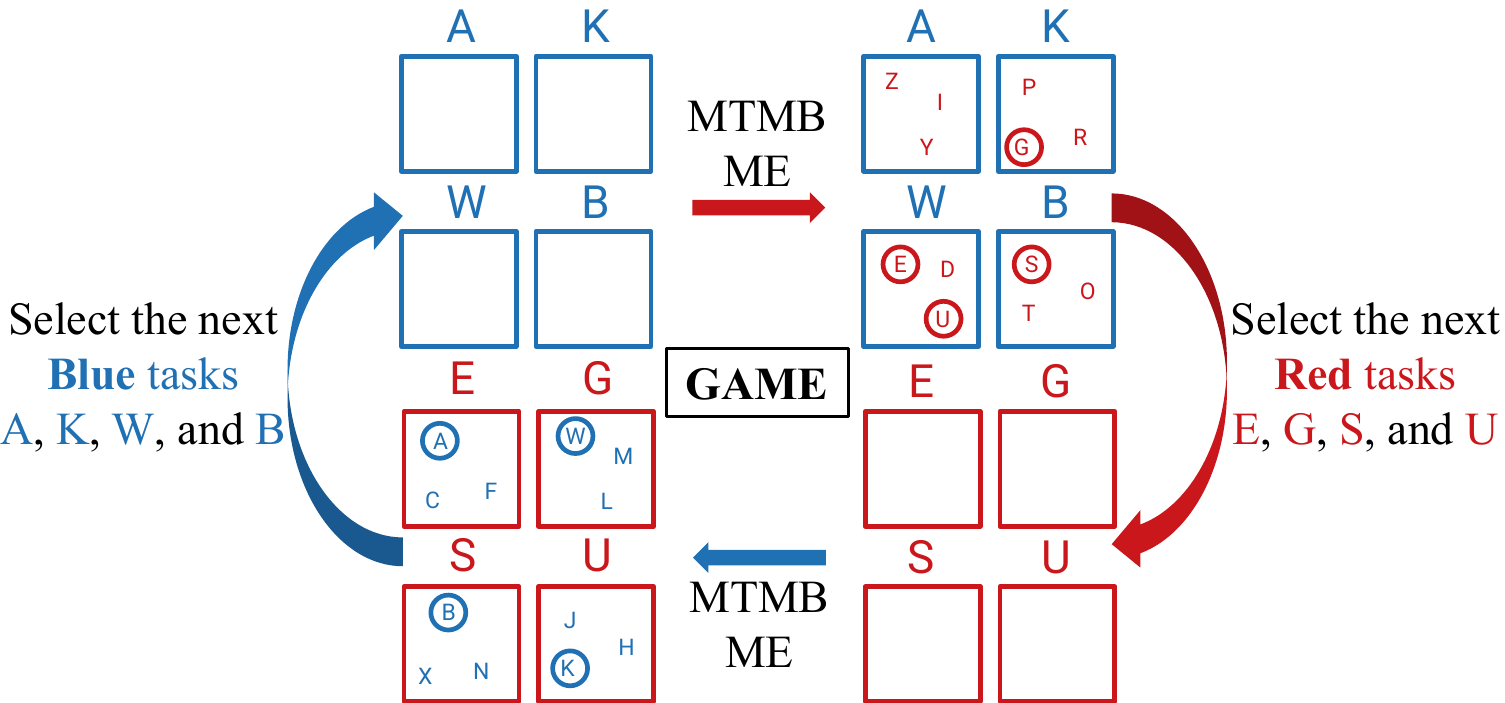}
    \vspace{-0.7cm}
    \caption{GAME is a coevolutionary QD algorithm that illuminates adversarial problems by alternating the execution of MTMB-ME~\citep{anne2023multi} on a set of tasks (i.e., fixed solutions from the opposing side) to encourage arms race dynamics. For example, in this illustration, the blue letters represent strategies a mouse uses to avoid a cat, and the red letters represent strategies a cat uses to catch a mouse. (Adapted from~\citep{anne2025generational}).}
    \label{fig:teaser}
    %\vspace{-0.5cm}
\end{figure}

Quality Diversity (QD)~\citep{pugh2016quality} is an evolutionary computation subfield that finds a set of diverse, high-quality solutions, in a process called illumination. It has been applied to domains such as robotics~\citep{cully2015robots}, video games~\citep{gravina2019procedural}, chemical synthesis~\citep{jiang2022artificial}, or aeronautics~\citep{brevault2024bayesian}. 

QD can be used to illuminate adversarial problems, for which it can be critical to identify all possible attack strategies, for example, to evaluate a system's current safety and defend accordingly. Examples of applications include: video games for automatic balancing of competitive games~\citep{fontaine2019mapping, fontaine2020covariance}, generalization evaluation of machine learning models~\citep{steckel2021illuminating, samvelyan2024multi}, and red teaming~\citep{ganguli2022red}, i.e., finding adversarial prompts that generate harmful content~\citep{Samvelyan2024RainbowTOA, wang2025quality, dang2025rainbowplus, nguyen2025diversifying}.

Those works illuminate only one side of the adversarial problem while fixing the other, thereby providing only partial illumination against the set of opposing problems picked by the experimenter. Generational Adversarial MAP-Elites (GAME)~\citep{anne2025generational} has recently been proposed as a general coevolutionary adversarial QD algorithm that illuminates both sides of an adversarial problem. GAME alternates the illumination of each side in a sequence of generations, using Multi-Task Multi-Behavior MAP-Elites (MTMB-ME)~\citep{anne2023multi} to illuminate one side against a fixed set of opposing solutions selected from the previous generation, called tasks (Fig.~\ref{fig:teaser}). For example, in a game of Cat-and-mouse, it corresponds to finding different strategies for the cat to catch the mouse and for the mouse to avoid the cat.  

The task selection mechanism drives the coevolutionary process and should select tasks that present greater diversity of challenges at each generation to broaden the illumination. The original GAME uses a behavioral criterion to select tasks; we argue that by disregarding the problem's adversarial aspect, it fails to provide the greatest illumination. In addition, we argue that classic quality and diversity metrics are not suited to adversarial problems. 

In this paper, we propose (1) two new task selection methods, \textbf{Ranking} and \textbf{Pareto}, that we compare against the original method (\textbf{Behavior}) and a random baseline (\textbf{Random}); (2) six metrics of adversarial quality and diversity; and (3) the evaluation on three adversarial problems: Pong, Cat-and-mouse, and Pursuers-and-evaders. 

The main contributions of this paper are: 
\begin{itemize}
    \item comparison of six metrics to evaluate the quality and diversity of a set of solutions on three adversarial problems;
    \item comparison of two new task selection mechanisms informed by a tournament and based on ranking and Pareto front optimality for GAME, showing that using adversarial information from a tournament results in higher quality and diversity than the original method. 
\end{itemize}

%Code, videos, and supplementary material are available at \url{https://anonymous.4open.science/r/TI-AQD}.

%%%%%%%%%%%%%%%%%%%%%%%%%%%%%%%%%%%%%%%%%%%%%%%%%%%%%%%%%%%%%%%%%%%%%%%%%%%%%%%%
%%%   Problem Statement
%%%%%%%%%%%%%%%%%%%%%%%%%%%%%%%%%%%%%%%%%%%%%%%%%%%%%%%%%%%%%%%%%%%%%%%%%%%%%%%%
\section{Problem Statement}
We define an adversarial QD problem as a tuple ($\mathcal{S}_{Red}$, $\mathcal{S}_{Blue}$, $\mathcal{F}$, $\mathcal{B}$) with $\mathcal{S}_{Red}$ and $\mathcal{S}_{Blue}$ the two opposing search spaces (they can be identical or different spaces), $\mathcal{F}$ is a fitness function defined by:
\begin{equation*}
    \mathcal{F}: \begin{array}{rcl}
S_{\text{Red}} \times S_{\text{Blue}} &\xrightarrow{}& [0,1]^2 \\
(s_{\text{red}}, s_{\text{blue}}) &\mapsto& (f_{\text{red}}, f_{\text{blue}})
\end{array}
\text{s.t. }  f_{\text{red}} + f_{\text{blue}} = 1 
\end{equation*}
and $\mathcal{B}$ is a behavior descriptor function defined by: 
\begin{equation*}
    \mathcal{B}: \begin{array}{rcl}
S_{\text{Red}} \times S_{\text{Blue}} &\xrightarrow{}& \mathbb{R}^m \\
(s_{\text{red}}, s_{\text{blue}}) &\mapsto& b
\end{array}.
\end{equation*} 
The problem is adversarial because the fitness of one side cannot increase without the other decreasing, and because the behavior descriptor depends on both sides. This raises multiple challenges: 
\begin{itemize}
    \item optimizing both search spaces simultaneously is challenging because an increase in fitness can result from either high-quality on one side or low-quality on the other;
    \item there is no intrinsic behavior descriptor for a solution.
\end{itemize}

This paper proposes various metrics for adversarial QD corresponding to different end goals for illumination. 

%%%%%%%%%%%%%%%%%%%%%%%%%%%%%%%%%%%%%%%%%%%%%%%%%%%%%%%%%%%%%%%%%%%%%%%%%%%%%%%%
%%%   Related Work
%%%%%%%%%%%%%%%%%%%%%%%%%%%%%%%%%%%%%%%%%%%%%%%%%%%%%%%%%%%%%%%%%%%%%%%%%%%%%%%%
\section{Related Work}

%%%%%%%%%%%%%%%%%%%%%%%%%%%%%%%%%%%%%%%%%%%%%%%%%%%%%%%%%%
%%  Quality Diversity
%%%%%%%%%%%%%%%%%%%%%%%%%%%%%%%%%%%%%%%%%%%%%%%%%%%%%%%%%%
\subsection{Quality Diversity}

A QD problem is defined by a fitness function to maximize and a behavior space to cover. Novelty Search with Local Competition (NSLC)~\citep{lehman2011evolving} and Multidimensional Archive of Phenotypic Elites (MAP-Elites)~\citep{mouret2015illuminating} are two of the main QD algorithms. MAP-Elites discretizes the behavior space into cells and keeps, for each cell, the solution with the highest fitness found so far, called the elite, forming an archive of high-quality, diverse solutions. At each iteration, MAP-Elites generates a new solution candidate from the archive's elites using a variation operator, evaluates it, and adds it to the archive if the new solution belongs to an empty cell (i.e., new behavior) or has a greater fitness than the current elite of its cell. 

Multi-Task MAP-Elites (MT-ME)~\citep{mouret2020quality} is a variant of MAP-Elites that tackles multi-task problems by simultaneously searching for the optimal solution for each task, where a task is defined by a specific behavior and fitness function. The intuition is that similar tasks should have similar solutions, enabling greater sample efficiency than optimizing each task individually. 

MTMB-ME~\citep{anne2023multi} is the QD extension of MT-ME. It simultaneously searches for multiple sets of diverse solutions -- one set per task -- where each set performs well against its dedicated task. GAME~\citep{anne2025generational} leverages MTMB-ME's ability to solve multi-task QD problems to search for a set of diverse, high-quality solutions specific to each fixed opposing solution of the current generation, i.e., the tasks.

%%%%%%%%%%%%%%%%%%%%%%%%%%%%%%%%%%%%%%%%%%%%%%%%%%%%%%%%%%
%%  Quality Diversity for Adversarial Problems
%%%%%%%%%%%%%%%%%%%%%%%%%%%%%%%%%%%%%%%%%%%%%%%%%%%%%%%%%%
\subsection{Quality Diversity for Adversarial Problems}

QD algorithms are increasingly applied to adversarial problems, for which it is important to illuminate the set of high-performing solutions. It has been applied in video games to illuminate possible decks in a card game~\citep{fontaine2019mapping} or policies for playing those decks~\citep{fontaine2020covariance}. It can also be used to assess procedural content generation; for example, \citep{steckel2021illuminating} use MAP-Elites to examine the levels generated by different GANs in the Lone Runner game. MADRID~\citep{samvelyan2024multi} applies MAP-Elites to find adversarial scenarios and evaluate the generalization capacities of a reinforcement learning (RL) agent in the Google Research Football environment.

In recent years, QD has been highlighted for its efficiency in red teaming with numerous algorithms, such as Rainbow Teaming~\citep{Samvelyan2024RainbowTOA}, Quality-Diversity Red-Teaming~\citep{wang2025quality}, and RAINBOWPLUS~\citep{dang2025rainbowplus}. \citep{nguyen2025diversifying} also uses MAP-Elites to illuminate the space of possible adversarial attacks on a text-to-image generative model. 

Those methods address sample-efficiency challenges specific to LLMs, but compared with GAME, they optimize only one side of the adversarial problem. Due to the difficulty of maintaining an adversarial coevolutionary process~\citep{ficici1998challenges}, few methods attempt to search both sides of the adversarial problem. We can identify two types of such methods: self-play, which originates in RL, and adversarial coevolution, which originates in evolutionary computation.  

%%%%%%%%%%%%%%%%%%%%%%%%%%%%%%%%%%%%%%%%%%%%%%%%%%%%%%%%%%
%%   Self-Play
%%%%%%%%%%%%%%%%%%%%%%%%%%%%%%%%%%%%%%%%%%%%%%%%%%%%%%%%%%
\subsection{Self-Play}

Self-play~\citep{zhang2024survey} is an RL method that consists of training the agent against itself or previous versions of itself. It has enabled surpassing human performance in multiple adversarial domains, such as Chess~\citep{silver2017mastering}, GO~\citep{silver2017mastering}, Dota 2~\citep{berner2019dota}, and StarCraft~\citep{vinyals2019grandmaster}. \citep{baker2019emergent} uses adversarial self-play to train RL agents in a 3D multi-agent hide-and-seek game. They show that it enables the successive emergence of multiple strategies on each side, thereby showcasing an artificial arms-race dynamic. Digital Red Queen (DRQ)~\citep{kumar2026digital} is a self-play algorithm that leverages LLMs to evolve assembly scripts that compete for survival in a memory-bound virtual machine. DRQ sequentially evolves a solution that competes against all previous elite solutions, leveraging MAP-Elites at each generation to preserve diversity. DRQ is interested in an adversarial environment where a solution competes against multiple others to survive, while GAME is, in a sense, more general, as it can be applied to any adversarial problem, but is currently restricted to only two-opponent problems. 

While all those agents benefit from training against a diverse set of opponents, self-play seeks the most robust solutions rather than all possible ways to beat a given opponent. 

%%%%%%%%%%%%%%%%%%%%%%%%%%%%%%%%%%%%%%%%%%%%%%%%%%%%%%%%%%
%%   Adversarial Coevolution
%%%%%%%%%%%%%%%%%%%%%%%%%%%%%%%%%%%%%%%%%%%%%%%%%%%%%%%%%%
\subsection{Adversarial Coevolution}

Adversarial coevolution is one of the main challenges of artificial life~\citep{bedau2000open, dorin2024artificial}. One of its goals is to create an open-ended artificial process by following a virtuous circle of arms race dynamics~\citep{dawkins1979arms}. 

Some adversarial coevolution algorithms focus on coevolving RL agents on one side and the environments and tasks they must solve on the other~\citep{wang2019poet, wang2020enhanced, chigot2022coevolution, faldoromni}. The idea is to generate a curriculum as an open-ended sequence of environments that are neither too easy nor too hard, which enables the discovery of necessary stepping stones. Enhanced-POET~\citep{wang2020enhanced} introduces the Performance of All Transferred Agents – Environment Comparison (PATA-EC) metric of adversarial diversity to select interesting environments. It runs a round-robin tournament to collect fitness scores for all agents across all environments and uses the resulting ranking vector to compare environments. The idea is that diverse environments should pose diverse challenges, i.e., yield different rankings of the current agents. In this paper, we draw inspiration from PATA-EC to propose a task selection mechanism for GAME and a general metric of adversarial diversity that is not specific to agent-environment problems. 

Compared with GAME, these works seek a robust agent by generating diverse training environments but do not explicitly aim to illuminate all possible agent behaviors or environments.

%%%%%%%%%%%%%%%%%%%%%%%%%%%%%%%%%%%%%%%%%%%%%%%%%%%%%%%%%%
%%   Adversarial Coevolution
%%%%%%%%%%%%%%%%%%%%%%%%%%%%%%%%%%%%%%%%%%%%%%%%%%%%%%%%%%
\subsection{Adversarial Coevolutionary QD}

Few works tackle the adversarial coevolutionary QD problem. \citep{dharna2024quality} uses a self-play QD algorithm to coevolve Python scripts using LLMs in a pursuer-and-evader game (similar to Cat-and-mouse presented in this paper). An LLM estimates the genotypic diversity of scripts by comparing their embedding vectors, and uses an NSLC-type archive to store the different solutions. 

GAME~\citep{anne2025generational} is a recent adversarial coevolutionary QD algorithm that leverages MTMB-ME to coevolve both sides of an adversarial problem. It is a general QD algorithm applicable to any two-sided adversarial problem (symmetric or asymmetric). GAME sequentially iterates the illumination of each side by selecting a set of elites from the previous generation and fixing them as tasks that the current generation competes against. In its original version, those tasks are selected based on behavioral diversity and quality criteria. We argue that while being computationally cheap (as it directly reuses the evaluation performed by MTMB-ME), it disregards the adversarial aspect of the problem. In this paper, we propose and compare two new task selection mechanisms that make informed decisions based on a tournament. 

Another limitation of current adversarial QD is the difficulty of applying classic QD metrics of quality and diversity, such as Behavior Coverage, max Fitness, and QD-Score, to adversarial problems, since both behaviors and fitness depend on the opposing side. Taking the maximum fitness of one side can easily correspond to selecting the worst solution on the opposing side, providing little information about the quality of the solutions. Behavior Coverage can still be used to compare two runs on the same problem, but it doesn't really answer the question: "Do I have many different solutions for both sides?" because the diversity could arise from one side rather than both. There is a need for specific metrics of adversarial QD. This paper proposes six of them. 

%\vspace{-0.3cm}
\begin{algorithm} 
    \small
    \caption{GAME\\
    \textbf{Inputs}: \textcolor{red}{Red} search space \textcolor{red}{$\mathcal{S}_{Red}$}, \textcolor{blue}{Blue} search space \textcolor{blue}{$\mathcal{S}_{Blue}$} \\
    \textbf{Parameters}: $N_{gen}$, $N_{task}$, $\operatorname{Tasks\_Selection}$ 
    }\label{alg:main_alg}
\begin{algorithmic}[1]
\State{} $Tasks \leftarrow$ Sample $N_{task}$ random \textcolor{blue}{Blue} solutions \Comment{\textbf{(a)}}

\State{} $\mathrm{B} = \emptyset$ \Comment{For storing bootstrapping evaluations}
\For{$gen\_id=1$ \texttt{in} $N_{gen}$}
    \State{} S $\leftarrow$ \textcolor{red}{$\mathcal{S}_{Red}$} if $gen\_id$ is odd else \textcolor{blue}{$\mathcal{S}_{Blue}$}
    \State{} $\mathrm{A} \leftarrow \operatorname{MTMB-ME}(Tasks, S, \mathrm{B})$ \Comment{\textbf{(b-f) - Alg.~\ref{alg:MTMB-ME}}}
    \State{} $Tasks, \mathrm{B} \leftarrow \operatorname{Tasks\_Selection}(\mathrm{A}, Tasks) $ \Comment{\textbf{(g) - Alg.~\ref{alg:behavior_TS} and \ref{alg:ranking_TS}}}
\EndFor{}
\State{} \Return{} $Tasks$
\end{algorithmic}
\end{algorithm}
%\vspace{-0.3cm}

%%%%%%%%%%%%%%%%%%%%%%%%%%%%%%%%%%%%%%%%%%%%%%%%%%%%%%%%%%%%%%%%%%%%%%%%%%%%%%%%
%%%   Method
%%%%%%%%%%%%%%%%%%%%%%%%%%%%%%%%%%%%%%%%%%%%%%%%%%%%%%%%%%%%%%%%%%%%%%%%%%%%%%%%
\section{Method}

\begin{figure*}
    \centering
    \includegraphics[width=\linewidth]{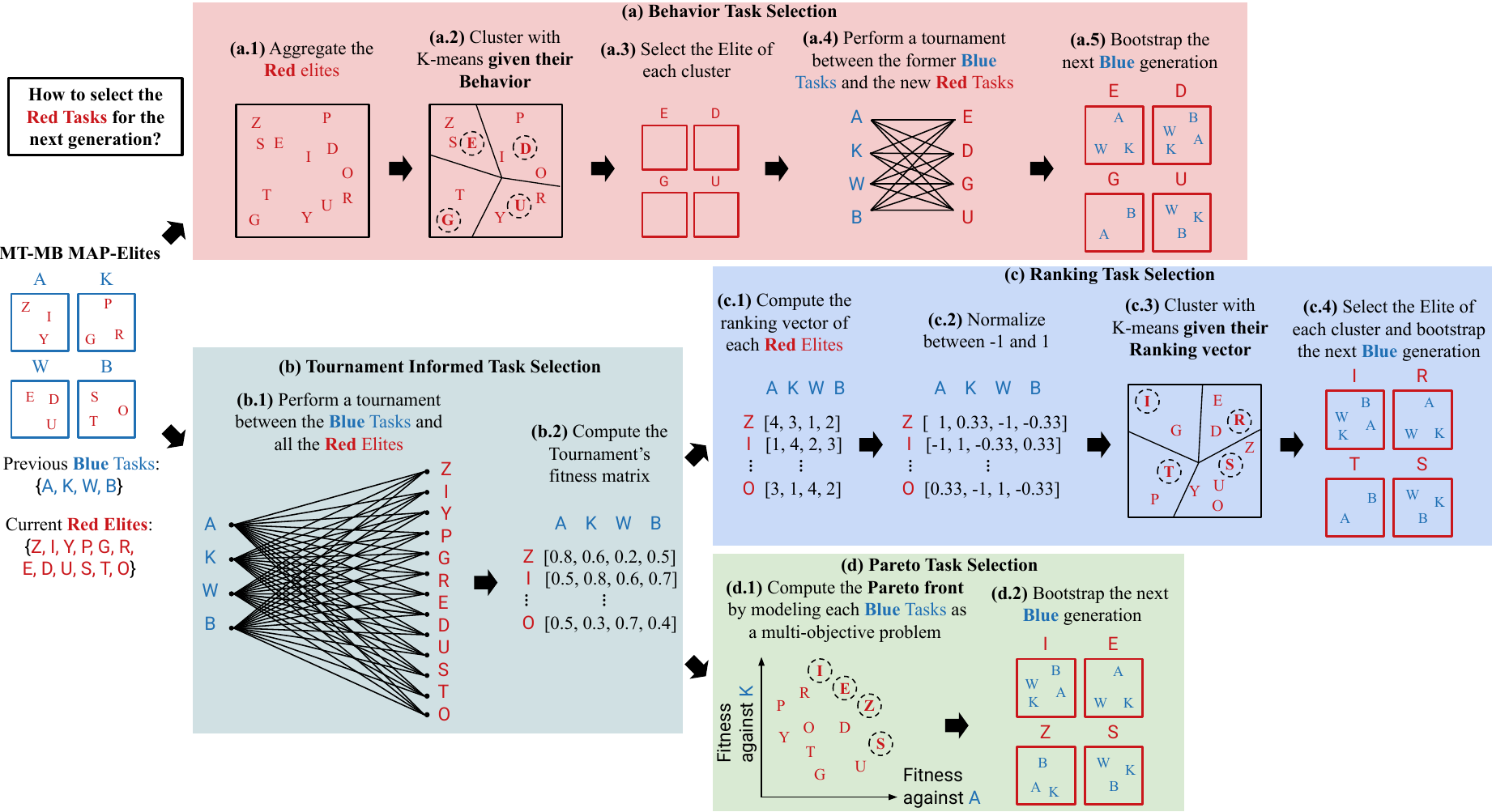}
    \vspace{-0.7cm}
    \caption{To illuminate an adversarial problem, GAME selects elites from the previous generation and sets them as opposing tasks for the next generation, which should represent a diversity of challenges. (a) In its original version, GAME selects tasks using a behavioral criterion. (b) In this paper, we propose two new methods that are informed by a tournament between the previous tasks and all elites: (c) \textbf{Ranking}, which selects a set of solutions that present a diversity of challenges, the idea being that different challenges should creates different rankings for the opposing side, and (d) \textbf{Pareto}, which selects a Pareto Front of solutions by considering the current tasks as a multi-objective problem.}
    \label{fig:method}
    \vspace{-0.2cm}
\end{figure*}

%%%%%%%%%%%%%%%%%%%%%%%%%%%%%%%%%%%%%%%%%%%%%%%%%%%%%%%%%%
%%   
%%%%%%%%%%%%%%%%%%%%%%%%%%%%%%%%%%%%%%%%%%%%%%%%%%%%%%%%%%
\subsection{Generational Adversarial MAP-Elites}

%%%%%%%%%%%%%%%%%%%%%%%%%%%%%%%%%%%%%%%%%%
% GAME algorithm 
%%%%%%%%%%%%%%%%%%%%%%%%%%%%%%%%%%%%%%%%%%

GAME proceeds as follows:
\begin{itemize}[leftmargin=*,align=left,widest={(c)}]
    \item[\textbf{(a)}] randomly sample $N_{task}$ Blue solutions as tasks;
    
\noindent \textbf{For $N_{gen}$ generations:}
    \begin{itemize}[leftmargin=*]
        \item[\textbf{(b)}] initialize a multi-task multi-behavior growing archive with $N_{cell}$ for each task;
        
        \noindent \textbf{For $N_{budget}$ evaluations:}
        \begin{itemize}[leftmargin=*]
            \item[\textbf{(c)}] select a task at random;
            \item[\textbf{(d)}] randomly picked elites from the whole archive to generate a candidate solution $s$ using a variation operator;
            \item[\textbf{(e)}] evaluate $s$ against the $task$ to collect the fitness $f$ and behavior $b$;
            \item[\textbf{(f)}] update the task's archive (Alg.~\ref{alg:growing_archive}): (1) if the new behavior $b$ is farther from all the current cells' centroids than the closest pair of centroids, it is added to the archive and one of the two centroids from the pair is removed; (2) else if the new fitness $f$ is greater than the corresponding cell's fitness, then the new solution becomes the elite of its cell;
        \end{itemize}
        \item[\textbf{(g)}] select $N_{task}$ elites from the whole archive (of size $N_{cell}\cdot N_{task}$) to be the next generation of tasks (comparing different such methods is one of the contributions of this paper and is presented in the next section); 
        \item[\textbf{(h)}] bootstrap the next generation with the evaluations resulting from a tournament between the new and previous tasks.
    \end{itemize}
\end{itemize}

Algorithms~\ref{alg:main_alg}-\ref{alg:growing_archive} detail GAME's implementation. We follow the original implementation, which uses an unstructured archive (i.e., growing a CVT discretization of the behavior space over the iterations). It provides a general way to build an archive by specifying only the number of cells and not boundaries or a distance threshold (which can be challenging to set in high-dimensional spaces). The appendix Sec.~\ref{apx:DNS} presents a comparison with a variant of Ranking that uses DNS~\citep{bahlous2025dominated} instead of a growing archive.

%%%%%%%%%%%%%%%%%%%%%%%%%%%%%%%%%%%%%%%%%%
% MTMB-ME algorithm
%%%%%%%%%%%%%%%%%%%%%%%%%%%%%%%%%%%%%%%%%%
\begin{algorithm} 
    \small
    \caption{MTMB-ME with growing archive\\ 
    \textbf{Inputs}: $Tasks$, search space $S$, bootstrap set $\mathrm{B}$\\
    \textbf{Parameters}: Number of evaluations $N_{budget}$, Number of initial random search $N_{init}$, evaluation function $\operatorname{Evaluate}$, variation operator $\operatorname{Variation}$, VEM 
    }\label{alg:MTMB-ME}
\begin{algorithmic}[1]
\State{} $\mathrm{A} \leftarrow$ Initialize $N_{task}$ growing archives \Comment{\textbf{(b)}}
\For{$(task, s, f, b)$ \texttt{in} $\mathrm{B}$} \Comment{\textbf{(g)} - Bootstrapping}
   \State{}  $\mathrm{A}[task] \leftarrow \operatorname{Update}(\mathrm{A}[task], s, f, b)$ \Comment{\textbf{Alg.~\ref{alg:growing_archive}}}
\EndFor{}
\For{$i=1$ \texttt{to} $N_{budget}$} \Comment{Main loop}
    \State{} $task \leftarrow$  Select a task at random from $Tasks$ \Comment{\textbf{(c)}}
    \If{ $\mathrm{A}$ has fewer than $N_{init}$ elites} 
        \State{} $s \leftarrow$ Sample a random solution from $S$ \Comment{\textbf{(d)}}
    \Else{}
        \State{} $s \leftarrow \operatorname{Variation}(\mathrm{A})$  \Comment{\textbf{(d)}}
    \EndIf{}
    \State{} $f, b \leftarrow \operatorname{Evaluate}(s, task)$ \Comment{where Evaluate uses $\mathcal{F}$ and $\mathcal{B}$ \textbf{(e)}}
    \State{}  $\mathrm{A} \leftarrow \operatorname{Update}(\mathrm{A}[task], s, f, b)$ \Comment{\textbf{(f) - Alg.~\ref{alg:growing_archive}}}
\EndFor{}
\State{} \Return{} Archives
\end{algorithmic}
\end{algorithm}

%%%%%%%%%%%%%%%%%%%%%%%%%%%%%%%%%%%%%%%%%%
% Growing archive
%%%%%%%%%%%%%%%%%%%%%%%%%%%%%%%%%%%%%%%%%%
\begin{algorithm} 
    \small
    \caption{Growing unstructured archive update\\ 
    \textbf{Inputs}: archive $A$, solution $s$, fitness $f$, behavior $b$\\
    \textbf{Parameters}: max archive size $N_{cell}$, distance function $\operatorname{dist}$
    }\label{alg:growing_archive}
\begin{algorithmic}[1] 
\State{} $(C, E, E_{backup}) = A$  \Comment{Centroids, Elites, and Backup Elites}
\If{$\operatorname{size}(C) < N_{cell}$} \Comment{Add a new cell}
    \State{} $i = \operatorname{size}(E)$
    \State{} $C[i] = b$ 
    \State{} $E[i] \leftarrow (s, f, b)$
    \State{} $E_{backup}[i] \leftarrow [(s, f, b)]$  
\Else{} \Comment{Check behavior and fitness}
    \State{} $distances = \{\operatorname{dist}(C[i], C[j])\}_{0 \le i < j < N_{cell}}$
    \State{} $d_{min} = \min(distances)$
    \State{} $d = \min\{\operatorname{dist}(b, C[i])\}_{0 \le i < N_{cell}}$
    \State{} $id = \operatorname{find\_cell}(C, b)$ \Comment{Closest centroid's index}
    \If{ $d > d_{min}$} \Comment{New enough behavior = growth}
        \State{} $j, k \leftarrow \operatorname{argmin}(distances)$  
        \State{} $d_j \leftarrow \min\{\operatorname{dist}(C[j], C[i])\}_{0 \le i \ne j < N_{cell} }$
        \State{} $d_k \leftarrow \min\{\operatorname{dist}(C[k], C[i])\}_{0 \le i \ne k < N_{cell} }$
        \State{} $k \leftarrow j $ if $d_j < d_k$ else $k$ 
        \State{} $C[k] \leftarrow b$
        \State{} $E[k] \leftarrow (s, f, b)$
        \State{} $E_{backup}[k] \leftarrow [(s, f, b)]$
        \For{$i=0$ to $N_{cell}-1$} \Comment{Check and repair holes}
            \If{$\operatorname{find\_cell}(C, E[i].b) \ne i$}
                \State{} $E[i] \leftarrow E_{backup}[i]$
            \EndIf{}
        \EndFor{}
    \ElsIf{f $>$ E[$id$].f} \Comment{Better fitness}
        \State{} $E[id] \leftarrow (s, f, b)$
        \State{} $E_{backup}[id].\operatorname{append}((s, f, b))$
    \EndIf{}
\EndIf{}
\State{} \Return{} $(C, E, E_{backup})$
\end{algorithmic}
\end{algorithm}

%%%%%%%%%%%%%%%%%%%%%%%%%%%%%%%%%%%%%%%%%%%%%%%%%%%%%%%%%%
%%   
%%%%%%%%%%%%%%%%%%%%%%%%%%%%%%%%%%%%%%%%%%%%%%%%%%%%%%%%%%
\subsection{How to select the generation of tasks?}

The task selection guides coevolution by proposing diverse and challenging opponents from the previous generation. We present the original method, \textbf{Behavior}, a baseline, \textbf{Random}, and the two new methods: \textbf{Ranking} and \textbf{Pareto}. Fig.~\ref{fig:method} illustrates: (a) \textbf{Behavior}, (b) the task selection tournament, (c) \textbf{Ranking}, and (d) \textbf{Pareto}.

\subsubsection{Behavior}
\label{sec:method_behavior_ts}
The original GAME selects tasks based on a behavioral quality and diversity criterion (Alg.~\ref{alg:behavior_TS}). It (a.1) aggregates all elites using behavior collected from MTMB-ME's evaluations, (a.2) recomputes an archive as if they were from the same behavior space using $N_{task}$ cells, and (a.3) selects the elite of each cell, ignoring that they were evaluated on different tasks.

This raises multiple issues: (1) the behavior aggregation does not take into account that the behavior is task dependent, and (2) taking the elite of each new cell may favor elites originating from ``easier'' tasks. One benefit of this method is that the task selection does not require additional computation, and the bootstrap tournament is of minimal size $N_{task}^2 << N_{budget}$.  

%%%%%%%%%%%%%%%%%%%%%%%%%%%%%%%%%%%%%%%%%%
% Behavior Task selection
%%%%%%%%%%%%%%%%%%%%%%%%%%%%%%%%%%%%%%%%%%
\begin{algorithm} 
    \small
    \caption{Behavior Task Selection\\
    \textbf{Inputs}: multi-task archive $\mathrm{A}$, previous tasks $Tasks_{old}$\\
    \textbf{Parameters}: $N_{task}$ 
    }\label{alg:behavior_TS}
\begin{algorithmic}[1]
\State{} $Behaviors$ $\leftarrow$ Aggregate $\mathrm{A}$'s elites' behaviors
\State{} Clusters $\leftarrow \operatorname{K-means}(Behaviors, k=N_{task})$  \Comment{Cluster the elites}
\State{} $Tasks \leftarrow \{\operatorname{Elite}(cluster)\}_{cluster \in Clusters}$ \Comment{Select the elite of each}
\State{} $\mathrm{B} \leftarrow$ $Tasks$ versus $Tasks_{old}$ \Comment{Tournament}

\State{} \Return{} $Tasks$, $\mathrm{B}$
\end{algorithmic}
\end{algorithm}

%%%%%%%%%%%%%%%%%%%%%%%%%%%%%%%%%%%%%%%%%%
% Ranking Task selection
%%%%%%%%%%%%%%%%%%%%%%%%%%%%%%%%%%%%%%%%%%
\begin{algorithm}[h]
    \small
    \caption{Ranking Task Selection\\
    \textbf{Inputs}: archive $\mathrm{A}$, previous $Tasks_{old}$\\
    \textbf{Parameters}: $N_{task}$
    }\label{alg:ranking_TS}
\begin{algorithmic}[1]
\State{} $Elites$ $\leftarrow$ Aggregate $\mathrm{A}$'s elites \Comment{$|Elites|=N_{task}\cdot N_{cell}$}
\State{} $\mathcal{T} \leftarrow$ $Elites$ versus $Tasks_{old}$ \Comment{(b.1) Tournament}
\State{} $\underline{f_e}=(\mathcal{T}[e \text{ vs t}].f)_{t \in Tasks_{old}}\in \mathbb{R}^{N_{task}}$ \Comment{(b.2) Fitness vector}
\State{} $\underline{r_e} = \operatorname{argsort(\operatorname{argsort(\underline{f_e})})}$ \Comment{(c.1) Ranking vector$^*$}  
\State{} $\underline{r_e} = \frac{2\cdot\underline{r_e}}{N_{task}-1}-1$ \Comment{(c.2) Normalize between -1 and 1} 
\State{} $Clusters \leftarrow \operatorname{K-means}(\{\underline{r_e}\}_{e \in Elites}, k=N_{task})$  \Comment{(c.3) Cluster}
\State{} $Tasks \leftarrow \{\operatorname{Elite}(cluster)_{cluster \in Clusters}\}$  \Comment{(c.4) Select elites}
\State{} $\mathrm{B} \leftarrow$ evaluations from $\mathcal{T}$ that includes elites from $Tasks$
\State{} \Return{} $Tasks$, $\mathrm{B}$
\end{algorithmic}
\footnotesize{$^*$ $\operatorname{argsort}^2 =$ inverse permutation of the sorted indices $=$ the ranking}
\end{algorithm}

\subsubsection{Random}
\label{sec:method_random_ts}
As a baseline, we compare with selecting elites at random, which also requires no additional evaluation. 

\subsubsection{Ranking}
\label{sec:method_ranking_ts}
We propose a selection mechanism inspired by PATA-EC that uses the previous tasks to estimate, via a tournament, the elites that will propose diverse challenges (Alg.~\ref{alg:ranking_TS}). 

It (b.1) performs a tournament between all the elites and the previous tasks (size $N_{task}^2\cdot N_{cell}$) to (b.2) collect the fitness vector. Then it (c.1) computes the ranking vector of the different tasks for each elite, (c.2) normalizes this ranking, (c.3) uses this as an adversarial behavior descriptor to cluster all the elites in $N_{task}$ cells, and (c.4) selects the elite of each cell, using the average fitness over all tasks as quality criteria. Note that all the elites have been evaluated against all tasks, so the comparison is fairer than with \textbf{Behavior}. Finally, for the bootstrapping, only the evaluations from the selected elites are used, meaning that most of the evaluations of the tournament (i.e., $N_{task}^2\cdot (N_{cell}-1)$) are not repurposed. 

\subsubsection{Pareto}
\label{sec:method_pareto_ts}
We can view the fitness vector from the tournament as a set of $N_{task}$ objectives to optimize. Searching for a diverse set of challenges corresponds to selecting the elites from the Pareto front of the multi-objective optimization. To do so, we use NSGA-III~\citep{deb2013evolutionary} to select $N_{task}$ elites from the successive Pareto fronts of the tournament's fitness vectors.  

%%%%%%%%%%%%%%%%%%%%%%%%%%%%%%%%%%%%%%%%%%%%%%%%%%%%%%%%%%
%%   
%%%%%%%%%%%%%%%%%%%%%%%%%%%%%%%%%%%%%%%%%%%%%%%%%%%%%%%%%%
\subsection{Metrics of adversarial QD}
We propose six metrics of quality and diversity for adversarial QD. They are all computed from an inter-variant tournament (which differs from the bootstrapping or the task selection tournaments), in which the last generation of tasks for each variant is selected and evaluated against one another. This has the advantage of evaluating all solutions against the same set of opponents at the cost of additional evaluations (i.e., $( N_{task}\cdot N_{rep} \cdot N_{variant})^2$). 

Without loss of generality, in the following formulation, we evaluate the quality and diversity of a set $S_{red}$ of solutions from one variant against the aggregate set $\underline{S}_{blue}$ of solutions from all variants. And use $f_{s_{red},s_{blue}}$ as the fitness of the solution $s_{red}\in S_{red}$ against the solution $s_{blue} \in \underline{S}_{blue}$.

%%%%%%%%%%%%%%%%%%%%%%%%%%%%%%%%%
%  Win rate    
%%%%%%%%%%%%%%%%%%%%%%%%%%%%%%%%%
\subsubsection{Win rate}
\textit{``Did it find a solution that wins against most opponents?''} As the adversarial fitness as the propriety $f_{\text{red}} + f_{\text{blue}} = 1$, we can set a winning threshold at $0.5$ and compute the win rate of each solution $s_{red}$ as $win\ rate(s_{red}) = \frac{1}{|\underline{S}_{blue}|}\underset{s_{blue}\in \underline{S}_{blue}}{\sum} \mathds{1}_{f_{s_{red},s_{blue}}>0.5}$, and define 
$$\textbf{Win rate}(S_{red})=\underset{s_{red}\in S_{red}}{\operatorname{max}}win\ rate(s_{red}).$$
This is a quality metric that can also be interpreted as an evaluation of the sides' balance. 

%%%%%%%%%%%%%%%%%%%%%%%%%%%%%%%%%
%  ELO score  
%%%%%%%%%%%%%%%%%%%%%%%%%%%%%%%%%
\subsubsection{ELO score}
\textit{``Did it find a solution that wins against strong opponents?''} We compute from the tournament's winnings the ELO score~\citep{elo1978rating} of each solution, which estimates the relative skill between solutions. To ease the comparison, we compute the ranking resulting from this ELO score among all solutions of the different compared variants, normalize between $0\%$ (lowest ELO score rank) and $100\%$ (highest ELO score rank), and pick the highest rank: 
$$\textbf{ELO score}(S_{red})=\underset{s_{red}\in S_{red}}{\operatorname{max}}ELO\_score\_rank(s_{red}).$$
Compared to \textbf{Win rate}, \textbf{ELO score} corrects potential imbalance of the opposing side, by giving more value to winning against a few strong opponents than winning against numerous weak opponents. It can be seen as a robust estimation of quality for comparing variants, but cannot be directly interpreted.  

%%%%%%%%%%%%%%%%%%%%%%%%%%%%%%%%%
%  Robustness   
%%%%%%%%%%%%%%%%%%%%%%%%%%%%%%%%%
\subsubsection{Robustness}
\textit{``Did it find a solution with no weakness?''} We define the robustness of a solution as its worst fitness: $$robustness(s_{red})=\underset{s_{blue}\in \underline{S}_{blue}}{\operatorname{min}} f_{s_{red}, s_{blue}}$$ and
$$\textbf{Robustness}(S_{red})=\underset{s_{red}\in S_{red}}{\operatorname{max}}robustness(s_{red}).$$
This is a different quality metric that emphasizes having no weaknesses rather than being good against good solutions (\textbf{ELO score}) or against most solutions (\textbf{Win rate}). 

%%%%%%%%%%%%%%%%%%%%%%%%%%%%%%%%%
%  Coverage  
%%%%%%%%%%%%%%%%%%%%%%%%%%%%%%%%%
\subsubsection{Coverage}
\textit{``Does the set propose a diverse set of challenges?''} Inspired by \textbf{Ranking}, we propose a metric of adversarial diversity that uses the normalized ranking vector of each solution as an adversarial behavioral descriptor of dimension the number of total solutions of the opposing sides ($N_{variant}\cdot N_{replication}\cdot N_{task}$). We cluster all the solutions $\underline{S}_{red}$ with K-means ($k=N_{task}$), and then count the percentage of cells filled by solutions from $S_{red}$ as a metric of adversarial coverage:
$$\textbf{Coverage}(S_{red})=\frac{|\{C_{s_{red}} \}_{s_{red} \in S_{red}}|}{|S_{red}|} \text{ where } C_{s_{red}} \text{ is } s_{red}\text{'s cluster}.$$ This is a metric of adversarial diversity that should be maximal if the set of solutions covers all possible challenges.

%%%%%%%%%%%%%%%%%%%%%%%%%%%%%%%%%
%  Expertise
%%%%%%%%%%%%%%%%%%%%%%%%%%%%%%%%%
\subsubsection{Expertise}
\textit{``Does the set contain counter-solutions?''} A diverse set of solutions should be diverse enough that for any opponent, there is at least one solution of the set that is strong against it. In other words, the set should contain a counter-solution for any opposing solution. To evaluate that, we compute the highest fitness obtained against each opposing solution and then pick the lowest of those, which corresponds to the worst counter-solution: 
$$\textbf{Expertise}(S_{red})=\underset{s_{blue}\in \underline{S}_{blue}}{\operatorname{min}}\underset{s_{red}\in S_{red}}{\operatorname{max}}f_{s_{red}, s_{blue}}.$$
This is a metric for adversarial QD, as it requires having covered all relevant strategies with high-quality solutions. 

%%%%%%%%%%%%%%%%%%%%%%%%%%%%%%%%%
%  AQD score
%%%%%%%%%%%%%%%%%%%%%%%%%%%%%%%%%
\subsubsection{Adversarial QD-Score}
\textit{``Does the set propose strictly different challenges?''} In contrast to \textbf{Expertise}, a diverse and high-quality set of solutions should require a large number of different opposing solutions for each of its solutions to lose. We thus compute the size of the smallest set of counter-solutions needed to make each solution lose at least once:
\begin{gather*}
\textbf{AQD score}(S_{red}) = \underset{S_{blue} \subseteq \underline{S}_{blue},}{\min} \left| S_{blue} \right|  \\
\text{ subject to } \, \forall s_{red} \in S_{red}, \exists s_{blue} \in S_{blue} \text{ such that } f_{s_{red}, s_{blue}} < 0.5 .\end{gather*}
%We use a linear programming solver to compute it. 

%%%%%%%%%%%%%%%%%%%%%%%%%%%%%%%%%%%%%%%%%%%%%%%%%%%%%%%%%%%%%%%%%%%%%%%%%%%%%%%%
%%%   Experiments  
%%%%%%%%%%%%%%%%%%%%%%%%%%%%%%%%%%%%%%%%%%%%%%%%%%%%%%%%%%%%%%%%%%%%%%%%%%%%%%%%
\section{Experiments}

\begin{figure}
    \centering
    \includegraphics[width=\linewidth]{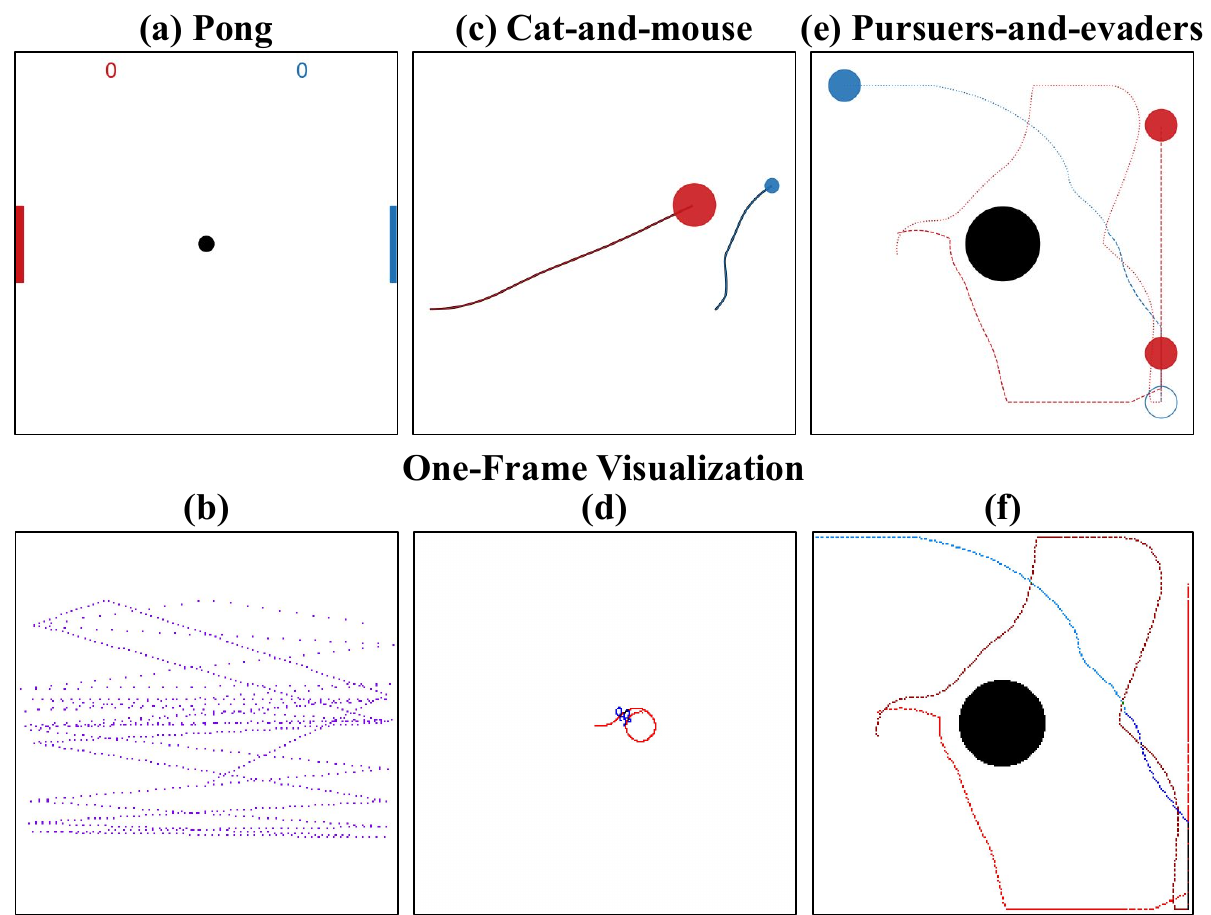}
    \vspace{-0.7cm}
    \caption{Pretty (top) and one-frame (bottom) visualization of the adversarial problems. (a-b) Pong, (c-d) Cat-and-mouse, and (e-f) Pursuers-and-evaders. The one-frame visualizations are from duels between the two solutions with the highest ELO score from the inter-variant tournament. Videos of those duels are available in the supplementary material.}
    \vspace{-0.5cm}
    \label{fig:environments}
\end{figure}

%%%%%%%%%%%%%%%%%%%%%%%%%%%%%%%%%%%%%%%%%%%%%%%%%%%%%%%%%%
%%   
%%%%%%%%%%%%%%%%%%%%%%%%%%%%%%%%%%%%%%%%%%%%%%%%%%%%%%%%%%
\subsection{Environments}
To evaluate both the task selection methods and the metrics of adversarial QD, we designed three increasingly complex adversarial problems. For all of them, the search space is the set of weights for a small MLP with hidden layers of sizes 32 and 16. The variation operator used is a mutation that applies Gaussian noise with standard deviation $0.1$ to $30\%$ of the weights. 

For the behavior space, we follow the original GAME and use a VEM (CLIP~\citep{radford2021learning}) that embeds a visualization of the adversarial problem (described in the following sections). All problems are implemented in JAX~\citep{jax2018github} to facilitate parallel evaluation on a GPU. 

%%%%%%%%%%%%%%%%%%%%%%%%%%%%%%%%%
%  Pong
%%%%%%%%%%%%%%%%%%%%%%%%%%%%%%%%%
\subsubsection{Pong}
We implemented the two-player Pong game (Fig.~\ref{fig:environments}.a), in which the ball's speed increases by 5\% after each successful rebound to increase difficulty. Each side, left and right, scores one point at each successful ball that the opponent does not rebound, after which the ball resets randomly at the center of the arena. 

The MLPs' input is the concatenation of the ball's position and velocity and the vertical positions of the paddles ($D=6$), each normalized to [-1, 1]. 
The MLPs' output is the vertical velocity of the paddle (output of tanh multiplied by the paddle velocity of 2\% of the arena height per time step). 
The fitness function is the ratio of points scored to the total points from both sides after 1000 steps. It is bounded between 0 and 1 (we set it to 0.5 if no points have been scored to represent a tie). The one-frame visualization shows the ball position at each time step, with colors changing after a point is scored (Fig.~\ref{fig:environments}.b). This adversarial problem is symmetrical. Ties can easily occur because either side scores no points (often) or both sides score the same number of points (rarely).

%%%%%%%%%%%%%%%%%%%%%%%%%%%%%%%%%
%  Cat-and-mouse
%%%%%%%%%%%%%%%%%%%%%%%%%%%%%%%%%
\subsubsection{Cat-and-mouse}
Cat-and-mouse (Fig.~\ref{fig:environments}.c) is an implementation with a neural network of the Homicidal Chauffeur problem~\citep{isaacs1999differential}. In this asymmetrical and adversarial problem, the cat, in red, is fast ($v_{cat} =2\ \mathrm{ms}^{-1}$ but cannot turn easily (i.e., its angular velocity is bounded so that the radius of the smallest full turn it can do is $r_{min}=1\ \mathrm{m}$) and is chasing a mouse, in blue, which is slower ($v_{cat} =1\ \mathrm{ms}^{-1}$) but is agile (i.e., it can change direction instantly). 

The MLPs' input is the concatenation of the cat's position and angular direction, and the mouse's position ($D=5$). 
The MLPs' output is the new angular direction (tanh output) multiplied by $\frac{v_{cat}}{r_{min}}=2\ \mathrm{rad\,s}^{-1}$ for the cat and $\pi$ for the mouse.  
The duel lasts until the cat catches the mouse (i.e., the distance between the two is lower than a threshold of $0.2\ \mathrm{m}$) or a maximum number of 500 timesteps is reached (i.e., $T_{max} = 5\ \mathrm{s}$). If the cat catches the mouse, the fitness is 1 minus the ratio of the time to catch the mouse, scaled between 0.5 and 1, so that 1 means catching the mouse at $t=0$ and 0.5 at $t=T_{max}$. If the cat does not catch the mouse, the fitness is the closest distance between the two, normalized by the initial distance and the catching threshold to be between 0 and 0.5, 0 meaning the cat was the closest at the start, and 0.5 that the cat was exactly at the distance threshold, so that the fitness function is continuous between the two modes. The one-frame visualization shows the positions of the cat (red) and the mouse (blue) throughout the duel (Fig.~\ref{fig:environments}.d). For vectorization, the duel continues for the maximum number of timesteps, but the fitness is computed early if the cat catches the mouse. 

%%%%%%%%%%%%%%%%%%%%%%%%%%%%%%%%%
%  Pursuers-and-evaders
%%%%%%%%%%%%%%%%%%%%%%%%%%%%%%%%%
\subsubsection{Pursuers-and-evaders}
In Pursuers-and-evaders (Fig.~\ref{fig:environments}.e), two pursuers, in red, must catch two evaders, in blue, i.e., getting closer than the catching threshold of $0.15\ \mathrm{m}$ before the end ($T_{max}=5s$). The pursuers and evaders have the same velocity ($v=1\ \mathrm{m\,s^{-1}}$), but the arena is bounded, and a central disc blocks the motion. %This environment is more complex due to the interactions among the four players. 

The pursuers share the same MLP, which takes as input the concatenation of its absolute position, the relative position of the two evaders, the relative position of the other pursuer, the id (1 for the first pursuer or -1 for the second), and the truth value if the first evader was caught and if the second evader was caught ($D=11$). Similarly, the evaders share the same MLP with a corresponding input.
The output of each MLP is the direction (tanh output scaled by $\pi$).
Similarly to Cat-and-mouse, the fitness is split into different modes: (a) if the two evaders are caught at time $T_1$ and $T_2$, it is the ratio $1-\frac{T_1 + T_2}{2T_{max}}$ normalized between  0.5 and 1 so that, 1 means that the pursuers caught them at $t=0\ \mathrm{s}$ and 0.5 that they caught both of them at $t=T_{max}$; (b) if only one evader is caught, it is the normalized closest distance to the remaining evader (between the initial distance and the capture threshold) between 0.25 and 0.5; and (c) if no evaders has been caught, it is the sum of the closest distance to the two evaders scaled between 0 and 0.25 by the initial position and the capture threshold. 

The one-frame visualization shows the positions of the pursuers (in red and brown) and the evaders (in light blue and dark blue) throughout the duel (Fig.~\ref{fig:environments}.f). This adversarial problem is also asymmetrical, and even with identical velocities, it should favor the pursuers due to the enclosed arena, a factor offset by the MLP's limited size and the short duration. 

\begin{table*}[]
\caption{Variants comparison of the different metrics in Pong}
\label{tab:pong}
\vspace{-0.4cm}
\footnotesize
\begin{NiceTabular}{@{}>{\raggedleft\arraybackslash}m{2.2cm}cccccccc@{}}
\CodeBefore
  \rowcolor{gray!15}{4}
  \rowcolor{gray!15}{6}
  \rowcolor{gray!15}{8}
\Body
\toprule
 & \multicolumn{2}{c}{\textbf{Ranking}} & \multicolumn{2}{c}{\textbf{Pareto}} & \multicolumn{2}{c}{\textbf{Behavior}} & \multicolumn{2}{c}{\textbf{Random}} \\
 & Left & Right & Left & Right & Left & Right & Left & Right \\
\midrule
\makecell[r]{\textbf{Win rate}} & \textbf{\makecell{59.8\% \\ {\footnotesize {[}58.4\%, 60.8\%{]}}}} & \textbf{\makecell{64.1\% \\ {\footnotesize {[}62.5\%, 65.9\%{]}}}} & \textbf{\makecell{59.9\% \\ {\footnotesize {[}58.0\%, 61.3\%{]}}}} & \makecell{61.7\% \\ {\footnotesize {[}60.8\%, 63.1\%{]}}} & \makecell{53.8\% \\ {\footnotesize {[}52.7\%, 54.8\%{]}}} & \makecell{57.5\% \\ {\footnotesize {[}55.9\%, 60.2\%{]}}} & \makecell{54.0\% \\ {\footnotesize {[}49.8\%, 55.1\%{]}}} & \makecell{58.3\% \\ {\footnotesize {[}57.0\%, 59.7\%{]}}} \\
\makecell[r]{\textbf{ELO score}} & \textbf{\makecell{99.3\% \\ {\footnotesize {[}98.5\%, 99.6\%{]}}}} & \textbf{\makecell{98.5\% \\ {\footnotesize {[}97.3\%, 99.7\%{]}}}} & \textbf{\makecell{99.3\% \\ {\footnotesize {[}98.2\%, 99.8\%{]}}}} & \textbf{\makecell{99.3\% \\ {\footnotesize {[}98.4\%, 99.6\%{]}}}} & \makecell{92.9\% \\ {\footnotesize {[}90.7\%, 94.9\%{]}}} & \makecell{87.9\% \\ {\footnotesize {[}82.7\%, 95.1\%{]}}} & \makecell{93.3\% \\ {\footnotesize {[}83.1\%, 95.4\%{]}}} & \makecell{82.2\% \\ {\footnotesize {[}72.4\%, 87.1\%{]}}} \\
\makecell[r]{\textbf{Robustness}} & \textbf{\makecell{0.00 \\ {\footnotesize {[}0.00, 0.00{]}}}} & \textbf{\makecell{0.00 \\ {\footnotesize {[}0.00, 0.00{]}}}} & \textbf{\makecell{0.00 \\ {\footnotesize {[}0.00, 0.00{]}}}} & \textbf{\makecell{0.00 \\ {\footnotesize {[}0.00, 0.00{]}}}} & \textbf{\makecell{0.00 \\ {\footnotesize {[}0.00, 0.00{]}}}} & \textbf{\makecell{0.00 \\ {\footnotesize {[}0.00, 0.00{]}}}} & \textbf{\makecell{0.00 \\ {\footnotesize {[}0.00, 0.00{]}}}} & \textbf{\makecell{0.00 \\ {\footnotesize {[}0.00, 0.00{]}}}} \\
\makecell[r]{\textbf{Coverage}} & \makecell{40.0\% \\ {\footnotesize {[}34.0\%, 42.0\%{]}}} & \textbf{\makecell{49.0\% \\ {\footnotesize {[}44.0\%, 54.0\%{]}}}} & \makecell{46.0\% \\ {\footnotesize {[}42.0\%, 48.0\%{]}}} & \textbf{\makecell{46.0\% \\ {\footnotesize {[}42.0\%, 48.0\%{]}}}} & \textbf{\makecell{54.0\% \\ {\footnotesize {[}50.0\%, 56.0\%{]}}}} & \textbf{\makecell{44.0\% \\ {\footnotesize {[}43.5\%, 51.0\%{]}}}} & \textbf{\makecell{52.0\% \\ {\footnotesize {[}49.5\%, 54.5\%{]}}}} & \textbf{\makecell{46.0\% \\ {\footnotesize {[}44.0\%, 50.0\%{]}}}} \\
\makecell[r]{\textbf{Expertise}} & \textbf{\makecell{0.50 \\ {\footnotesize {[}0.50, 0.50{]}}}} & \textbf{\makecell{0.50 \\ {\footnotesize {[}0.50, 0.50{]}}}} & \textbf{\makecell{0.50 \\ {\footnotesize {[}0.50, 0.50{]}}}} & \textbf{\makecell{0.50 \\ {\footnotesize {[}0.50, 0.50{]}}}} & \textbf{\makecell{0.50 \\ {\footnotesize {[}0.50, 0.50{]}}}} & \textbf{\makecell{0.50 \\ {\footnotesize {[}0.50, 0.50{]}}}} & \textbf{\makecell{0.50 \\ {\footnotesize {[}0.50, 0.50{]}}}} & \textbf{\makecell{0.50 \\ {\footnotesize {[}0.50, 0.50{]}}}} \\
\makecell[r]{\textbf{AQD score}} & \textbf{\makecell{3 \\ {\footnotesize {[}3, 3{]}}}} & \textbf{\makecell{3 \\ {\footnotesize {[}3, 3.2{]}}}} & \textbf{\makecell{3 \\ {\footnotesize {[}3, 3{]}}}} & \textbf{\makecell{3 \\ {\footnotesize {[}3, 4{]}}}} & \makecell{2 \\ {\footnotesize {[}2, 2{]}}} & \makecell{2 \\ {\footnotesize {[}2, 2{]}}} & \makecell{2 \\ {\footnotesize {[}2, 2{]}}} & \makecell{2 \\ {\footnotesize {[}2, 2{]}}} \\
\bottomrule
\end{NiceTabular}
\vspace{-0.2cm}
\end{table*}

%%%%%%%%%%%%%%%%%%%%%%%%%%%%%%%%%%%%%
%   Cat-and-mouse
%%%%%%%%%%%%%%%%%%%%%%%%%%%%%%%%%%%%%
\begin{table*}[]
\caption{Variants comparison of the different metrics in Cat-and-mouse}
\label{tab:cat_and_mouse}
\vspace{-0.4cm}
\footnotesize
\begin{NiceTabular}{@{}>{\raggedleft\arraybackslash}m{2.2cm}cccccccc@{}}
\CodeBefore
  \rowcolor{gray!15}{4}
  \rowcolor{gray!15}{6}
  \rowcolor{gray!15}{8}
\Body
\toprule
 & \multicolumn{2}{c}{\textbf{Ranking}} & \multicolumn{2}{c}{\textbf{Pareto}} & \multicolumn{2}{c}{\textbf{Behavior}} & \multicolumn{2}{c}{\textbf{Random}} \\
 & Cat & Mouse & Cat & Mouse & Cat & Mouse & Cat & Mouse \\
\midrule
\makecell[r]{\textbf{Win rate}} & \textbf{\makecell{56.7\% \\ {\footnotesize {[}51.0\%, 60.6\%{]}}}} & \textbf{\makecell{87.4\% \\ {\footnotesize {[}86.5\%, 88.4\%{]}}}} & \textbf{\makecell{52.6\% \\ {\footnotesize {[}48.4\%, 59.3\%{]}}}} & \textbf{\makecell{87.7\% \\ {\footnotesize {[}87.3\%, 88.3\%{]}}}} & \makecell{42.0\% \\ {\footnotesize {[}36.4\%, 46.8\%{]}}} & \makecell{85.2\% \\ {\footnotesize {[}84.9\%, 85.7\%{]}}} & \makecell{39.5\% \\ {\footnotesize {[}35.1\%, 44.7\%{]}}} & \makecell{85.3\% \\ {\footnotesize {[}84.7\%, 85.7\%{]}}} \\
\makecell[r]{\textbf{ELO score}} & \textbf{\makecell{99.3\% \\ {\footnotesize {[}98.3\%, 99.7\%{]}}}} & \textbf{\makecell{98.8\% \\ {\footnotesize {[}97.3\%, 99.6\%{]}}}} & \textbf{\makecell{98.7\% \\ {\footnotesize {[}97.5\%, 99.6\%{]}}}} & \textbf{\makecell{99.1\% \\ {\footnotesize {[}98.7\%, 99.5\%{]}}}} & \makecell{94.1\% \\ {\footnotesize {[}89.1\%, 96.8\%{]}}} & \makecell{91.7\% \\ {\footnotesize {[}89.8\%, 94.5\%{]}}} & \makecell{92.1\% \\ {\footnotesize {[}87.1\%, 95.7\%{]}}} & \makecell{92.3\% \\ {\footnotesize {[}88.6\%, 94.2\%{]}}} \\
\makecell[r]{\textbf{Robustness}} & \textbf{\makecell{0.40 \\ {\footnotesize {[}0.39, 0.42{]}}}} & \makecell{0.17 \\ {\footnotesize {[}0.17, 0.18{]}}} & \makecell{0.38 \\ {\footnotesize {[}0.36, 0.40{]}}} & \makecell{0.18 \\ {\footnotesize {[}0.17, 0.18{]}}} & \makecell{0.35 \\ {\footnotesize {[}0.33, 0.36{]}}} & \textbf{\makecell{0.18 \\ {\footnotesize {[}0.18, 0.18{]}}}} & \makecell{0.35 \\ {\footnotesize {[}0.33, 0.36{]}}} & \textbf{\makecell{0.18 \\ {\footnotesize {[}0.18, 0.18{]}}}} \\
\makecell[r]{\textbf{Coverage}} & \textbf{\makecell{58.0\% \\ {\footnotesize {[}56.0\%, 62.5\%{]}}}} & \textbf{\makecell{62.0\% \\ {\footnotesize {[}58.0\%, 64.0\%{]}}}} & \makecell{56.0\% \\ {\footnotesize {[}50.0\%, 58.5\%{]}}} & \makecell{52.0\% \\ {\footnotesize {[}48.0\%, 56.5\%{]}}} & \makecell{48.0\% \\ {\footnotesize {[}46.0\%, 52.0\%{]}}} & \makecell{47.0\% \\ {\footnotesize {[}40.0\%, 48.5\%{]}}} & \makecell{55.0\% \\ {\footnotesize {[}51.5\%, 58.0\%{]}}} & \makecell{53.0\% \\ {\footnotesize {[}50.0\%, 58.0\%{]}}} \\
\makecell[r]{\textbf{Expertise}} & \textbf{\makecell{0.81 \\ {\footnotesize {[}0.75, 0.82{]}}}} & \textbf{\makecell{0.55 \\ {\footnotesize {[}0.55, 0.56{]}}}} & \makecell{0.60 \\ {\footnotesize {[}0.50, 0.64{]}}} & \makecell{0.55 \\ {\footnotesize {[}0.54, 0.55{]}}} & \makecell{0.47 \\ {\footnotesize {[}0.45, 0.48{]}}} & \makecell{0.53 \\ {\footnotesize {[}0.52, 0.54{]}}} & \makecell{0.48 \\ {\footnotesize {[}0.48, 0.49{]}}} & \makecell{0.54 \\ {\footnotesize {[}0.54, 0.54{]}}} \\
\makecell[r]{\textbf{AQD score}} & \textbf{\makecell{2 \\ {\footnotesize {[}2, 2{]}}}} & \textbf{\makecell{3 \\ {\footnotesize {[}3, 3{]}}}} & \makecell{2 \\ {\footnotesize {[}1, 2{]}}} & \makecell{3 \\ {\footnotesize {[}2.8, 3{]}}} & \makecell{1 \\ {\footnotesize {[}1, 1{]}}} & \makecell{2 \\ {\footnotesize {[}2, 2{]}}} & \makecell{1 \\ {\footnotesize {[}1, 1{]}}} & \makecell{2 \\ {\footnotesize {[}2, 2{]}}} \\
\bottomrule
\end{NiceTabular}
\vspace{-0.1cm}
\end{table*}

%%%%%%%%%%%%%%%%%%%%%%%%%%%%%%%%%%%%%
%   Pursuers-and-evaders
%%%%%%%%%%%%%%%%%%%%%%%%%%%%%%%%%%%%%
\begin{table*}[]
\caption{Variants comparison of the different metrics in Pursuers-and-evaders}
\label{tab:pursuers_and_evaders}
\vspace{-0.4cm}
\footnotesize
\begin{NiceTabular}{@{}>{\raggedleft\arraybackslash}m{2.2cm}cccccccc@{}}
\CodeBefore
  \rowcolor{gray!15}{4}
  \rowcolor{gray!15}{6}
  \rowcolor{gray!15}{8}
\Body
\toprule
 & \multicolumn{2}{c}{\textbf{Ranking}} & \multicolumn{2}{c}{\textbf{Pareto}} & \multicolumn{2}{c}{\textbf{Behavior}} & \multicolumn{2}{c}{\textbf{Random}} \\
 & Pursuers & Evaders & Pursuers & Evaders & Pursuers & Evaders & Pursuers & Evaders \\
\midrule
\makecell[r]{\textbf{Win rate}} & \textbf{\makecell{48.0\% \\ {\footnotesize {[}46.0\%, 50.1\%{]}}}} & \textbf{\makecell{91.9\% \\ {\footnotesize {[}91.0\%, 92.4\%{]}}}} & \makecell{45.2\% \\ {\footnotesize {[}42.1\%, 46.0\%{]}}} & \textbf{\makecell{90.9\% \\ {\footnotesize {[}89.7\%, 92.0\%{]}}}} & \makecell{37.5\% \\ {\footnotesize {[}36.2\%, 41.2\%{]}}} & \makecell{90.6\% \\ {\footnotesize {[}89.0\%, 91.1\%{]}}} & \makecell{41.5\% \\ {\footnotesize {[}37.7\%, 43.1\%{]}}} & \makecell{89.1\% \\ {\footnotesize {[}88.3\%, 90.2\%{]}}} \\
\makecell[r]{\textbf{ELO score}} & \textbf{\makecell{99.5\% \\ {\footnotesize {[}98.8\%, 99.8\%{]}}}} & \textbf{\makecell{99.3\% \\ {\footnotesize {[}98.5\%, 99.7\%{]}}}} & \makecell{98.5\% \\ {\footnotesize {[}96.5\%, 98.8\%{]}}} & \textbf{\makecell{98.2\% \\ {\footnotesize {[}95.9\%, 99.4\%{]}}}} & \makecell{89.1\% \\ {\footnotesize {[}85.8\%, 95.4\%{]}}} & \makecell{97.7\% \\ {\footnotesize {[}94.3\%, 98.5\%{]}}} & \makecell{95.7\% \\ {\footnotesize {[}89.5\%, 97.4\%{]}}} & \makecell{94.7\% \\ {\footnotesize {[}92.4\%, 97.0\%{]}}} \\
\makecell[r]{\textbf{Robustness}} & \textbf{\makecell{0.14 \\ {\footnotesize {[}0.13, 0.15{]}}}} & \makecell{0.12 \\ {\footnotesize {[}0.12, 0.12{]}}} & \makecell{0.13 \\ {\footnotesize {[}0.12, 0.13{]}}} & \makecell{0.12 \\ {\footnotesize {[}0.12, 0.13{]}}} & \textbf{\makecell{0.13 \\ {\footnotesize {[}0.12, 0.13{]}}}} & \textbf{\makecell{0.13 \\ {\footnotesize {[}0.12, 0.13{]}}}} & \textbf{\makecell{0.13 \\ {\footnotesize {[}0.12, 0.14{]}}}} & \makecell{0.12 \\ {\footnotesize {[}0.12, 0.12{]}}} \\
\makecell[r]{\textbf{Coverage}} & \makecell{52.0\% \\ {\footnotesize {[}47.5\%, 56.0\%{]}}} & \textbf{\makecell{55.0\% \\ {\footnotesize {[}50.0\%, 58.0\%{]}}}} & \makecell{44.0\% \\ {\footnotesize {[}40.0\%, 48.5\%{]}}} & \makecell{42.0\% \\ {\footnotesize {[}37.5\%, 46.0\%{]}}} & \textbf{\makecell{54.0\% \\ {\footnotesize {[}50.0\%, 58.0\%{]}}}} & \makecell{49.0\% \\ {\footnotesize {[}47.5\%, 56.0\%{]}}} & \textbf{\makecell{55.0\% \\ {\footnotesize {[}54.0\%, 60.0\%{]}}}} & \textbf{\makecell{56.0\% \\ {\footnotesize {[}52.0\%, 60.0\%{]}}}} \\
\makecell[r]{\textbf{Expertise}} & \textbf{\makecell{0.58 \\ {\footnotesize {[}0.49, 0.76{]}}}} & \textbf{\makecell{0.76 \\ {\footnotesize {[}0.59, 0.77{]}}}} & \makecell{0.48 \\ {\footnotesize {[}0.47, 0.49{]}}} & \makecell{0.58 \\ {\footnotesize {[}0.56, 0.61{]}}} & \makecell{0.49 \\ {\footnotesize {[}0.48, 0.49{]}}} & \makecell{0.56 \\ {\footnotesize {[}0.55, 0.62{]}}} & \makecell{0.49 \\ {\footnotesize {[}0.48, 0.49{]}}} & \makecell{0.60 \\ {\footnotesize {[}0.55, 0.76{]}}} \\
\makecell[r]{\textbf{AQD score}} & \textbf{\makecell{1.5 \\ {\footnotesize {[}1, 2{]}}}} & \textbf{\makecell{3 \\ {\footnotesize {[}3, 3{]}}}} & \makecell{1 \\ {\footnotesize {[}1, 1{]}}} & \textbf{\makecell{3 \\ {\footnotesize {[}3, 3{]}}}} & \makecell{1 \\ {\footnotesize {[}1, 1{]}}} & \textbf{\makecell{3 \\ {\footnotesize {[}3, 3{]}}}} & \makecell{1 \\ {\footnotesize {[}1, 1{]}}} & \textbf{\makecell{3 \\ {\footnotesize {[}3, 3{]}}}} \\
\bottomrule
\end{NiceTabular}
\vspace{-0.1cm}
\end{table*}

%%%%%%%%%%%%%%%%%%%%%%%%%%%%%%%%%%%%%%%%%%%%%%%%%%%%%%%%%%
%%   Results
%%%%%%%%%%%%%%%%%%%%%%%%%%%%%%%%%%%%%%%%%%%%%%%%%%%%%%%%%%
\subsection{Results}

We performed 20 replications of the four variants for each adversarial problem with $N_{gen}=10$, $N_{cell}=20$, $N_{task}=50$, and $N_{budget}=\num{100000}$. As \textbf{Ranking} and \textbf{Pareto} require additional evaluations to select the tasks, we increased $N_{budget}$ for \textbf{Random} and \textbf{Behavior} so that each variant has the same total number of evaluations, $N_{gen}\cdot (N_{budget}+N_{task}^2\cdot N_{cell})=1.5\,\text{M}$.

We present the median result (and first and third quartiles in brackets) over the 20 replications in Tab.~\ref{tab:pong} for Pong, Tab.~\ref{tab:cat_and_mouse} for Cat-and-mouse, and Tab.~\ref{tab:pursuers_and_evaders} for Pursuers-and-evaders. Bold highlights values that are not significantly different from the best-performing variant on each side, using the Holm-Bonferroni correction applied separately to each adversarial problem, with a p-value of 0.05. The box plots and swarm plots are presented in the appendix Sec.~\ref{apx:boxplots}. 

%%%%%%%%%%%%%%%%%%%%%%%%%%%%%%%%%%%%%%%%%%%%%
%  Variants Comparison
%%%%%%%%%%%%%%%%%%%%%%%%%%%%%%%%%%%%%%%%%%%%%
\subsubsection{Variants Comparison}
\textbf{Ranking} has the highest \textbf{Win rate}, \textbf{ELO score}, \textbf{Expertise}, and \textbf{AQD score} for each side across all three adversarial problems, the highest \textbf{Robustness} for one side across the two adversarial problems where it is not null, and the highest \textbf{Coverage} for 4 out of 6 comparisons. \textbf{Pareto} consistently shows lower or similar performance across all metrics compared to \textbf{Ranking}. \textbf{Behavior} and \textbf{Random} have similar performances across all metrics, with lower \textbf{Win rate} and ELO score on each side in all three adversarial problems, but the highest \textbf{Coverage} for Pong and for Pursuers-and-evaders. 

%%%%%%%%%%%%%%%%%%%%%%%%%%%%%%%%%%%%%%%%%%%%%
%  Environments takeaways
%%%%%%%%%%%%%%%%%%%%%%%%%%%%%%%%%%%%%%%%%%%%%
\subsubsection{Environments takeaways}

Pong is not an open-ended adversarial problem. The \textbf{Expertise} of 0.5 and \textbf{Robustness} of 0 for all replications of all variants shows that it is easy to find a counter-solution that creates a tie, but hard (or impossible) to do better. This may be caused by the current parameters, and a longer duration or faster ball acceleration could make it harder. 

In Cat-and-mouse, the highest mouse's \textbf{Win rate} is much higher than the cat's, indicating that, with the current parameters, it is easier for the mouse to escape the cat. However, \textbf{Expertise} being strictly above 0.5 for both sides indicates that it is always possible to win (given that we know the opposing solution). 

Pursuers-and-evaders has lower \textbf{Robustness} and \textbf{Expertise} than Cat-and-mouse, and an increased difference between the two sides, as the best pursuers have a \textbf{Win rate} below 50\% while the evaders have a \textbf{Win rate} above 90\%.

%%%%%%%%%%%%%%%%%%%%%%%%%%%%%%%%%%%%%%%%%%%%%
%  Metrics comparison
%%%%%%%%%%%%%%%%%%%%%%%%%%%%%%%%%%%%%%%%%%%%%
\subsubsection{Metrics comparison}

\textbf{Win rate} seems to be an effective quality metric that also provides insight into the balance between the two sides. However, one should keep in mind that the absolute value is sensitive to the dataset and may not provide a good estimate of the average win rate against any opponent.

\textbf{ELO score} appears to be correlated with \textbf{Win rate}, further validating its use as a quality metric for comparison. However, its normalized nature makes it relative to the current set of solutions and does not provide any illumination information. 

\textbf{Robustness}, which is always less than 0.5, indicates that the three adversarial problems are fair because no solution can always win (or such a solution has not been found). However, as shown in Pong, it is not usable if fitness is too sparse.  

\textbf{Coverage} is always challenging to compute in high-dimensional space. Its K-means discretization, which is between 40\% and 60\% for each problem, allows comparison of the variants' diversity but does not provide meaningful insight into their absolute value. In addition, its higher values do not correlate with greater ability to find counter-solutions, as shown in Pursuers-and-evaders. 

\textbf{Expertise} not only provides a useful quality and diversity metric but also useful insights, as it shows the possible quality if one were able to predict or quickly recognize the opposing solution.  

\textbf{AQD score} low values (with a maximum of 3) indicate that the three adversarial problems are not open-ended, i.e., do not contain a large variety of winning strategies. Still, the two tournament-informed task selection methods identified a set of solutions that require an additional counter-solution to defeat them, indicating their higher performance in adversarial QD. 

%%%%%%%%%%%%%%%%%%%%%%%%%%%%%%%%%%%
%  Validating Ranking's performance
%%%%%%%%%%%%%%%%%%%%%%%%%%%%%%%%%%%
\paragraph{Validating Ranking's performance} 
We performed an alternative inter-variant tournament using the same GAME executions, but with \textbf{Ranking} to select the final tasks from the other variants. The results, presented in the appendix Sec.~\ref{apx:ranking_check}, show that full \textbf{Ranking} still outperforms the other variants. This validates that its superior performance does not come solely from selecting the last set of tasks, but from sequential selection across generations.

%%%%%%%%%%%%%%%%%%%%%%%%%%%%%%%%%%%%%%%%%%%%%%%%%%%%%%%%%%%%%%%%%%%%%%%%%%%%%%%%
%%%   Discussion
%%%%%%%%%%%%%%%%%%%%%%%%%%%%%%%%%%%%%%%%%%%%%%%%%%%%%%%%%%%%%%%%%%%%%%%%%%%%%%%%
\section{Discussion and Future Work}

The result confirms that a tournament-informed task selection method yields higher adversarial quality and diversity than behavioral or random task selection. \textbf{Ranking}, being significantly better than \textbf{Pareto}, which may be slightly too quality-focused, further validates the intuition from PATA-EC. \textbf{Behavior} not being significantly better than \textbf{Random} also shows that behavioral quality and diversity from one evaluation does not generalize. 

The tournament-informed methods require evaluating each elite on each task (i.e., $N_{task}^2\cdot N_{cell}$ evaluations). Future work should focus on improving their sample efficiency, as it seems unlikely that the full tournament is necessary to select the next generation of tasks with the highest adversarial quality and diversity. 

Similarly, the proposed metrics require an expensive inter-variant tournament. Future work should focus on improving their sample efficiency. %For example, \textbf{ELO score}, designed for such a purpose, should be easy to estimate with significantly fewer evaluations. 
Bridging the metrics with those developed for zero-sum games in Game Theory~\citep{osborne2004introduction} could also provide guidance. 

The three adversarial problems lack open-endedness, either because of the problems themselves or because of the small neural networks used to solve them. Future work should validate whether using \textbf{Ranking} as the task selection allows for open-ended discovery of solutions in more complex adversarial problems.  

%The choice of metric depends on the end goal of the adversarial illumination, finding: a solution likely to win in general (\textbf{Win rate}) or against strong opponents (\textbf{ELO score}); a solution with no weaknesses (\textbf{Robustness}); all possible challenges, including the bad ones (\textbf{Coverage}); a set of counter-solutions (\textbf{Expertise}); a set of solutions that require diversity for the opposing side (\textbf{AQD score}).

%%%%%%%%%%%%%%%%%%%%%%%%%%%%%%%%%%%%%%%%%%%%%%%%%%%%%%%%%%%%%%%%%%%%%%%%%%%%%%%%
%%%   Conclusion
%%%%%%%%%%%%%%%%%%%%%%%%%%%%%%%%%%%%%%%%%%%%%%%%%%%%%%%%%%%%%%%%%%%%%%%%%%%%%%%%
\section{Conclusion}
GAME is a coevolutionary adversarial QD algorithm that alternates illumination between the two sides of an adversarial problem. It selects elites from the previous generation to serve as fixed opponents, thereby promoting arms-race dynamics. In this paper, we show that a tournament-informed task selection mechanism based on the ranking vector yields higher adversarial quality and diversity. We also present six metrics to evaluate various characteristics of adversarial illumination. Overall, our results suggest that a tournament-informed task selection is a promising direction for applying GAME to richer, more open-ended adversarial domains.  

%%%%%%%%%%%%%%%%%%%%%%%%%%%%%%%%%%%%%%%%%%%%%%%%%%%%%%%%%%%%%%%%%%%%%%%%%%%%%%%%
%%% Acknowledgments 
%%%%%%%%%%%%%%%%%%%%%%%%%%%%%%%%%%%%%%%%%%%%%%%%%%%%%%%%%%%%%%%%%%%%%%%%%%%%%%%%
\begin{acks}
Funded by the armasuisse S+T project F00-007.

\end{acks}

\bibliographystyle{ACM-Reference-Format}
\bibliography{biblio}

\newpage
\appendix
\renewcommand{\thefigure}{\thesection.\arabic{figure}}
\setcounter{figure}{0}
\renewcommand{\thetable}{\thesection.\arabic{table}}
\setcounter{table}{0}

%%%%%%%%%%%%%%%%%%%%%%%%%%%%%%%%%%%%%%%%%%%%%%%%%%%%%%%%%%
%%   Results
%%%%%%%%%%%%%%%%%%%%%%%%%%%%%%%%%%%%%%%%%%%%%%%%%%%%%%%%%%
\section{Main comparisons}
\label{apx:boxplots}

\begin{figure*}[!hbp]
    \centering
    \includegraphics[width=0.8\linewidth]{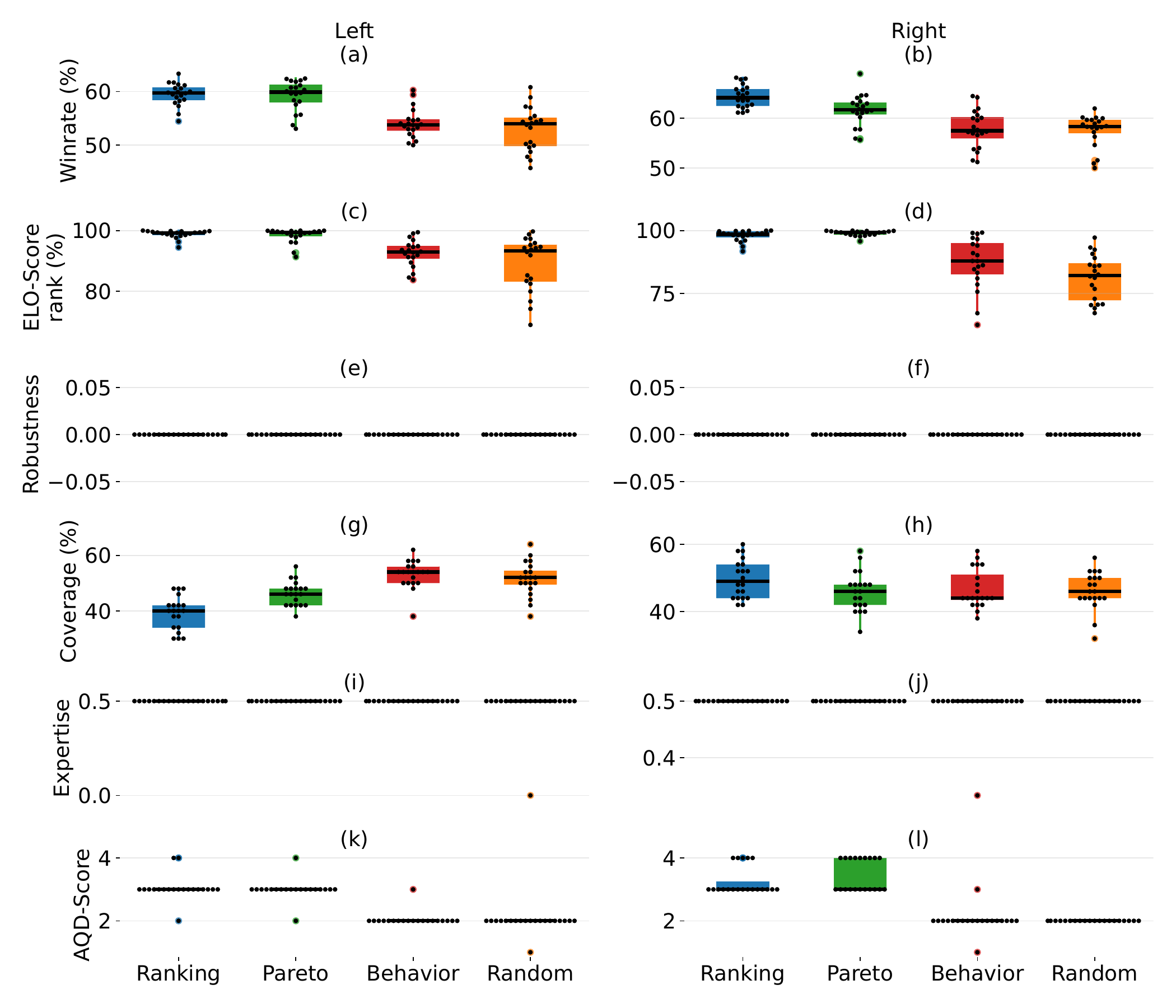}
    \caption{Pong box plots and swarm plots comparison of the different variants.}
    \label{fig:pong_comparison}
\end{figure*}

\begin{figure*}[!htbp]
    \centering
    \includegraphics[width=0.8\linewidth]{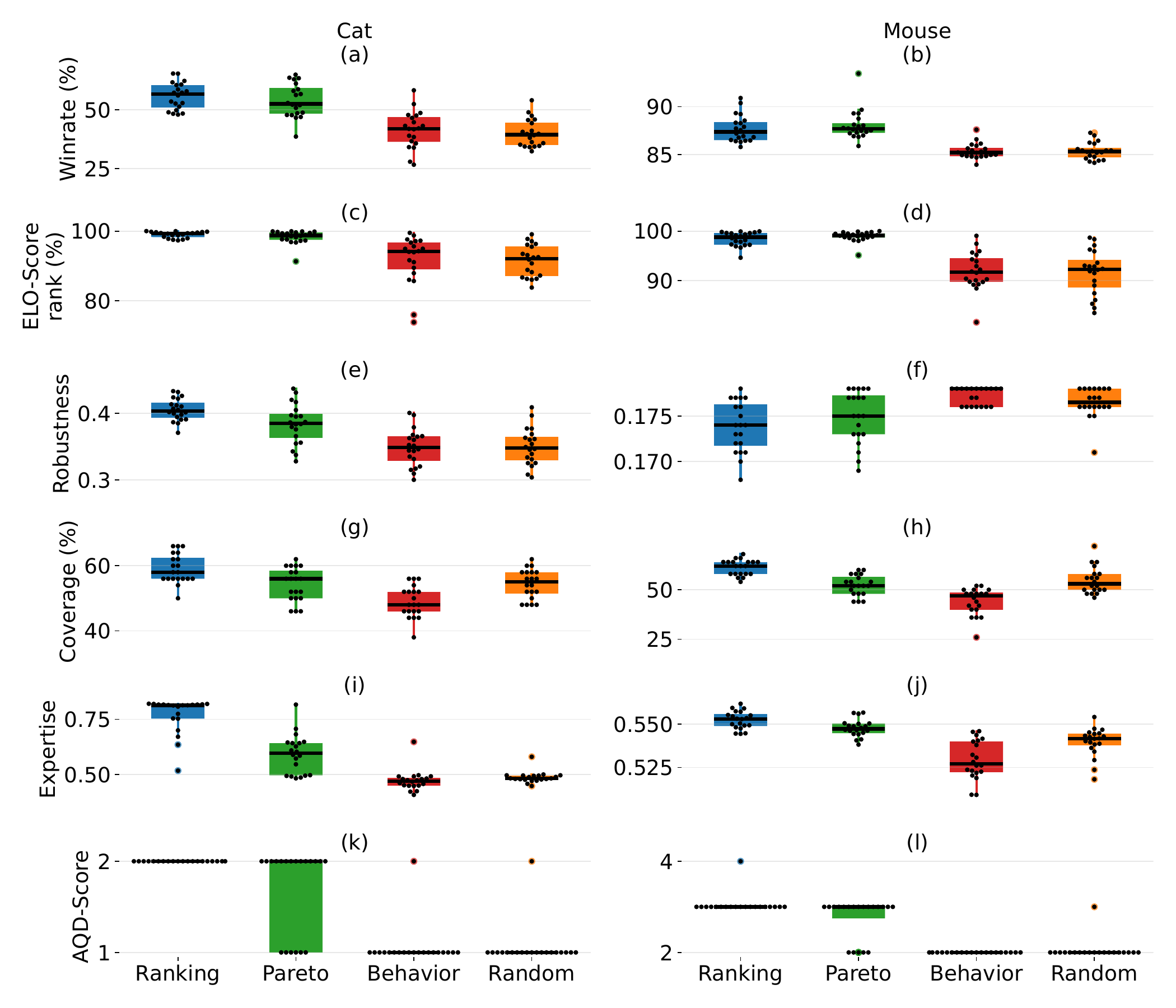}
    \caption{Cat-and-mouse box plots and swarm plots comparison of the different variants.}
    \label{fig:cat_and_mouse_comparison}
\end{figure*}

\begin{figure*}[!htbp]
    \centering
    \includegraphics[width=0.8\linewidth]{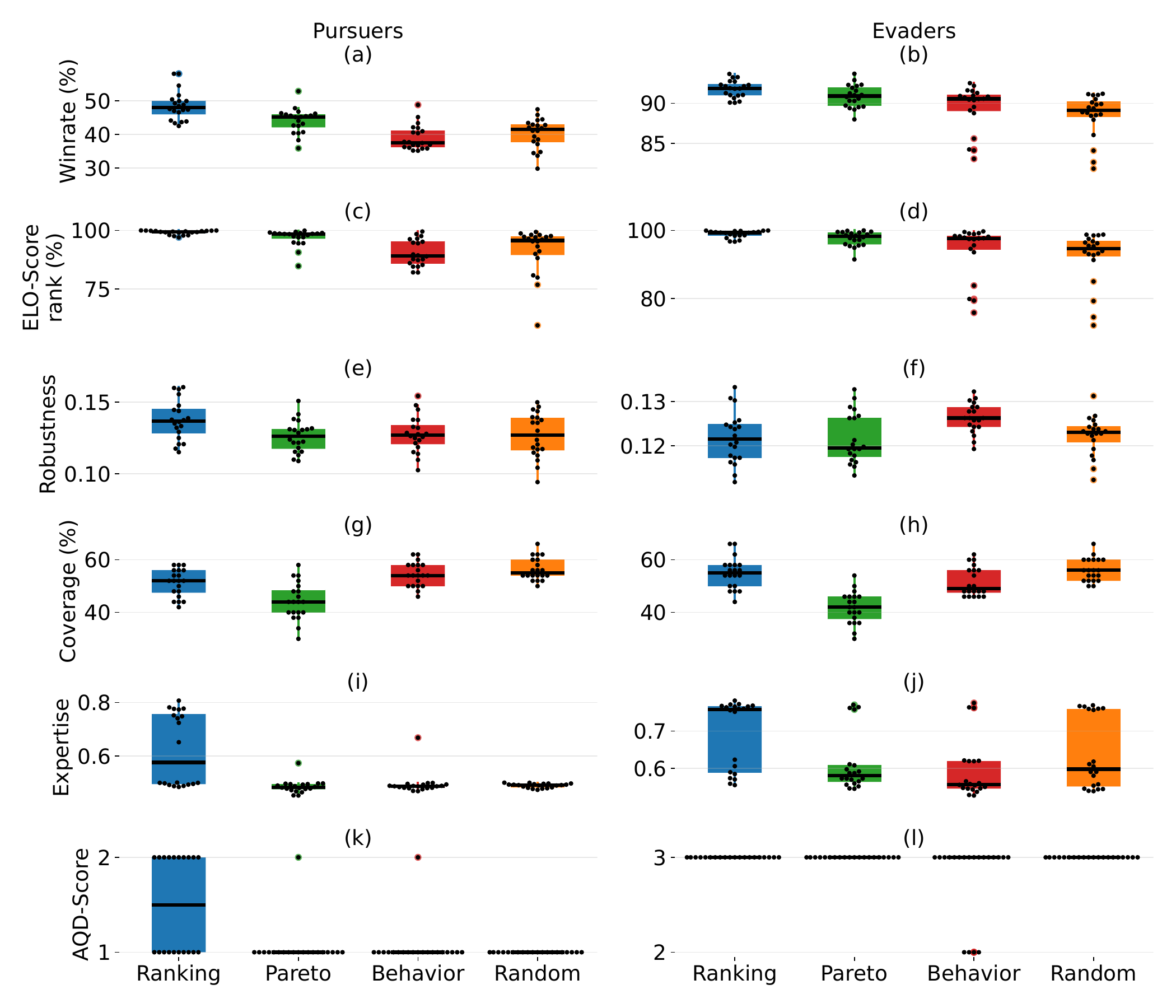}
    \caption{Pursuers-and-evaders box plots and swarm plots comparison of the different variants.}
    \label{fig:pursuers_and_evaders_comparison}
\end{figure*}

We present the box plots and swarm plots of the main experiment of the paper: Fig~\ref{fig:pong_comparison} for Pong; Fig~\ref{fig:cat_and_mouse_comparison} for Cat-and-mouse; and Fig~\ref{fig:pursuers_and_evaders_comparison} for Pursuers-and-evaders. 


\FloatBarrier 

%%%%%%%%%%%%%%%%%%%%%%%%%%%%%%%%%%%%%%%%%%%%%%%%%%%%%%%%%%
%%   Ranking sanity check
%%%%%%%%%%%%%%%%%%%%%%%%%%%%%%%%%%%%%%%%%%%%%%%%%%%%%%%%%%
\section{Validating Ranking's performance}
\label{apx:ranking_check}

To check that \textbf{Ranking}'s superior performance does not come only from selecting the last set of elites competing in the inter-variant tournament, but mainly from the sequential selection through the generations, we performed an alternative inter-variant tournament using the same executions of GAME as the main experiment, but selecting the competing tasks with \textbf{Ranking} from the final archives of each variant.

\begin{figure*}[!htbp]
    \centering
    \includegraphics[width=0.8\linewidth]{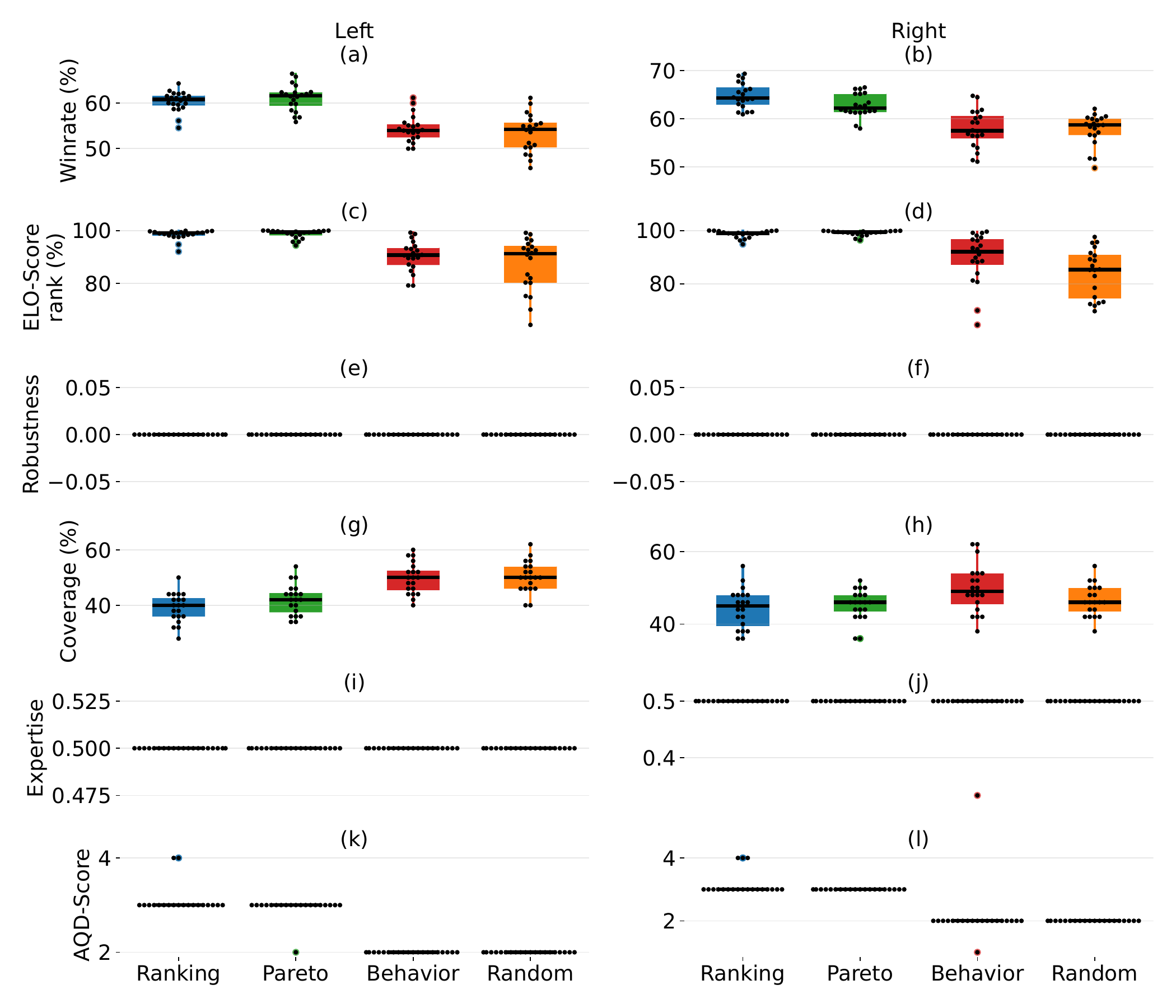}
    \caption{Pong comparison using \textbf{Ranking} to select the competing elites for each variant.}
    \label{fig:ranking_pong_comparison}
\end{figure*}

\begin{figure*}[ht]
    \centering
    \includegraphics[width=0.8\linewidth]{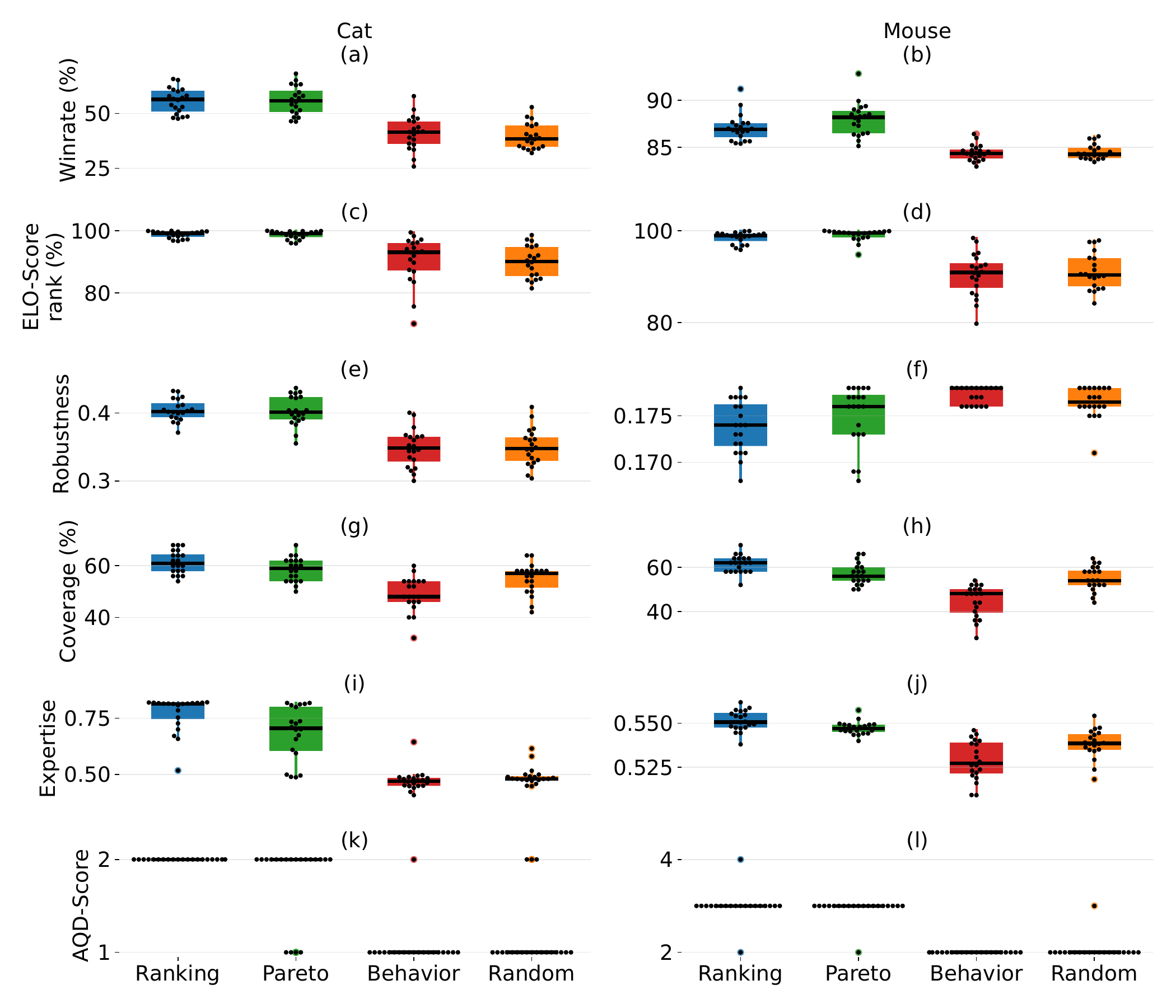}
    \caption{Cat-and-mouse comparison using \textbf{Ranking} to select the competing elites for each variant.}
    \label{fig:ranking_cat_and_mouse_comparison}
\end{figure*}

\begin{figure*}[ht]
    \centering
    \includegraphics[width=0.8\linewidth]{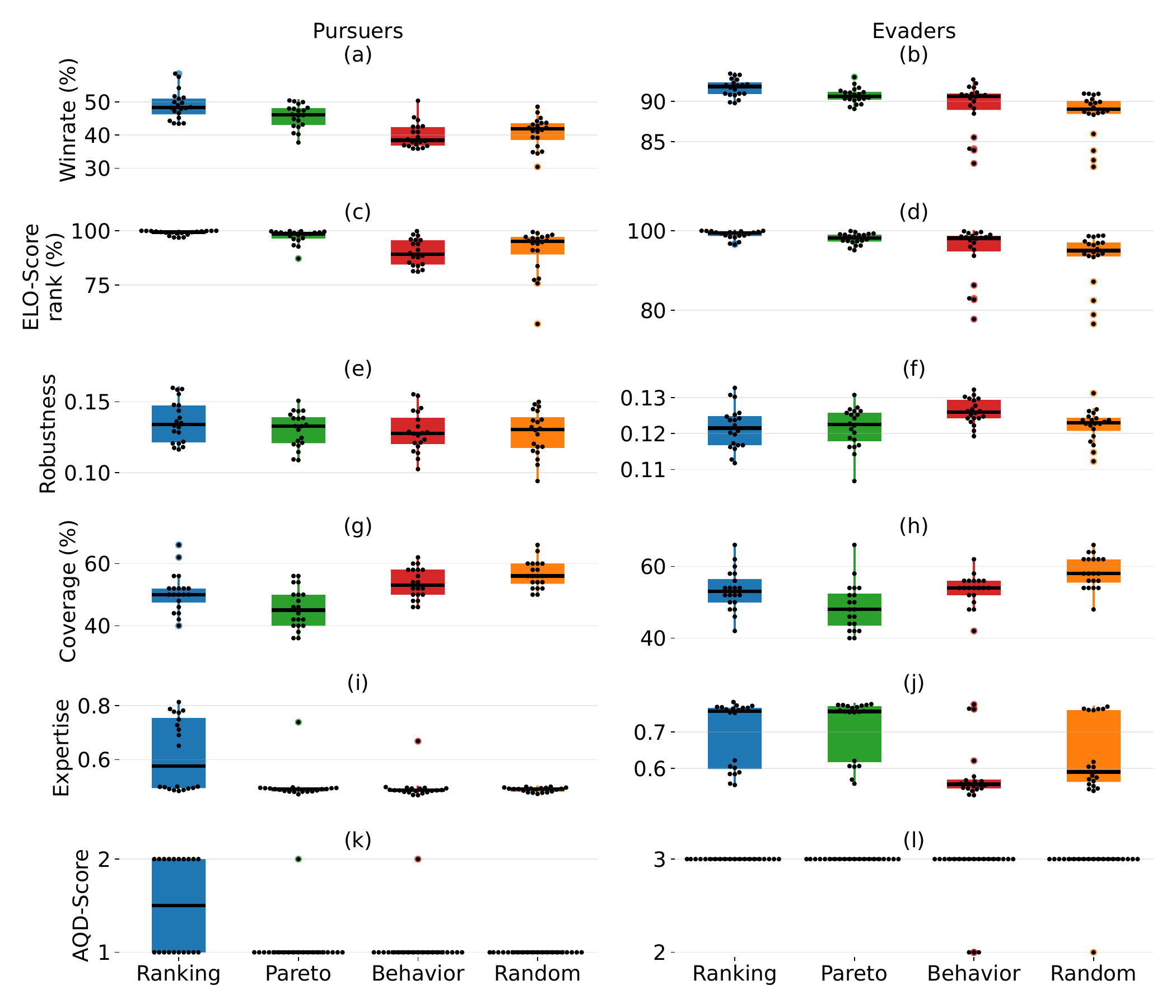}
    \caption{Pursuers-and-evaders comparison using \textbf{Ranking} to select the competing elites for each variant.}
    \label{fig:ranking_pursuers_and_evaders_comparison}
\end{figure*}

Fig.~\ref{fig:ranking_pong_comparison} and Tab.~\ref{tab:ranking_pong_comparison} present those results in Pong. 
Fig.~\ref{fig:ranking_cat_and_mouse_comparison} and Tab.~\ref{tab:ranking_cat_and_mouse_comparison} present those results in Cat-and-mouse.
Fig.~\ref{fig:ranking_pursuers_and_evaders_comparison} and Tab.~\ref{tab:ranking_pursuers_evaders_comparison} present those results in Pursuers-and-evaders. Bold values indicate values that are not significantly different from the best-performing variant, using the Holm-Bonferroni correction applied separately for each environment with a p-value of 0.05. 

\textbf{Ranking} still maintains higher performance across \textbf{Win rate}, \textbf{ELO-Score}, \textbf{Expertise}, and \textbf{AQD-Score}, while showing lower \textbf{Coverage} for Pong and Pursuers-and-Evaders, and lower \textbf{Robustness} for the mouse and evaders.

\begin{table*}[]
\caption{Pong comparison using \textbf{Ranking} to select the competing elites for each variant.}
\label{tab:ranking_pong_comparison}
\vspace{-0.4cm}
\footnotesize
\begin{NiceTabular}{@{}>{\raggedleft\arraybackslash}m{2.2cm}cccccccc@{}}
\CodeBefore
  \rowcolor{gray!15}{4}
  \rowcolor{gray!15}{6}
  \rowcolor{gray!15}{8}
\Body
\toprule
 & \multicolumn{2}{c}{\textbf{Ranking}} & \multicolumn{2}{c}{\textbf{Pareto}} & \multicolumn{2}{c}{\textbf{Behavior}} & \multicolumn{2}{c}{\textbf{Random}} \\
 & Left & Right & Left & Right & Left & Right & Left & Right \\
\midrule
\makecell[r]{\textbf{Win rate}} & \textbf{\makecell{60.8\% \\ {\footnotesize {[}59.4\%, 61.6\%{]}}}} & \textbf{\makecell{64.3\% \\ {\footnotesize {[}62.9\%, 66.5\%{]}}}} & \textbf{\makecell{61.6\% \\ {\footnotesize {[}59.4\%, 62.4\%{]}}}} & \makecell{62.2\% \\ {\footnotesize {[}61.4\%, 65.2\%{]}}} & \makecell{54.0\% \\ {\footnotesize {[}52.4\%, 55.3\%{]}}} & \makecell{57.5\% \\ {\footnotesize {[}55.9\%, 60.6\%{]}}} & \makecell{54.2\% \\ {\footnotesize {[}50.2\%, 55.7\%{]}}} & \makecell{58.7\% \\ {\footnotesize {[}56.6\%, 60.0\%{]}}} \\
\makecell[r]{\textbf{ELO Score}} & \textbf{\makecell{99.0\% \\ {\footnotesize {[}98.0\%, 99.5\%{]}}}} & \textbf{\makecell{99.0\% \\ {\footnotesize {[}98.3\%, 99.7\%{]}}}} & \textbf{\makecell{99.4\% \\ {\footnotesize {[}98.2\%, 99.8\%{]}}}} & \textbf{\makecell{99.4\% \\ {\footnotesize {[}98.7\%, 99.7\%{]}}}} & \makecell{90.7\% \\ {\footnotesize {[}87.0\%, 93.5\%{]}}} & \makecell{92.0\% \\ {\footnotesize {[}87.1\%, 96.8\%{]}}} & \makecell{91.2\% \\ {\footnotesize {[}80.3\%, 94.1\%{]}}} & \makecell{85.3\% \\ {\footnotesize {[}74.5\%, 90.9\%{]}}} \\
\makecell[r]{\textbf{Robustness}} & \textbf{\makecell{0.00 \\ {\footnotesize {[}0.00, 0.00{]}}}} & \textbf{\makecell{0.00 \\ {\footnotesize {[}0.00, 0.00{]}}}} & \textbf{\makecell{0.00 \\ {\footnotesize {[}0.00, 0.00{]}}}} & \textbf{\makecell{0.00 \\ {\footnotesize {[}0.00, 0.00{]}}}} & \textbf{\makecell{0.00 \\ {\footnotesize {[}0.00, 0.00{]}}}} & \textbf{\makecell{0.00 \\ {\footnotesize {[}0.00, 0.00{]}}}} & \textbf{\makecell{0.00 \\ {\footnotesize {[}0.00, 0.00{]}}}} & \textbf{\makecell{0.00 \\ {\footnotesize {[}0.00, 0.00{]}}}} \\
\makecell[r]{\textbf{Coverage}} & \makecell{40.0\% \\ {\footnotesize {[}36.0\%, 42.5\%{]}}} & \makecell{45.0\% \\ {\footnotesize {[}39.5\%, 48.0\%{]}}} & \makecell{42.0\% \\ {\footnotesize {[}37.5\%, 44.5\%{]}}} & \textbf{\makecell{46.0\% \\ {\footnotesize {[}43.5\%, 48.0\%{]}}}} & \textbf{\makecell{50.0\% \\ {\footnotesize {[}45.5\%, 52.5\%{]}}}} & \textbf{\makecell{49.0\% \\ {\footnotesize {[}45.5\%, 54.0\%{]}}}} & \textbf{\makecell{50.0\% \\ {\footnotesize {[}46.0\%, 54.0\%{]}}}} & \textbf{\makecell{46.0\% \\ {\footnotesize {[}43.5\%, 50.0\%{]}}}} \\
\makecell[r]{\textbf{Expertise}} & \textbf{\makecell{0.50 \\ {\footnotesize {[}0.50, 0.50{]}}}} & \textbf{\makecell{0.50 \\ {\footnotesize {[}0.50, 0.50{]}}}} & \textbf{\makecell{0.50 \\ {\footnotesize {[}0.50, 0.50{]}}}} & \textbf{\makecell{0.50 \\ {\footnotesize {[}0.50, 0.50{]}}}} & \textbf{\makecell{0.50 \\ {\footnotesize {[}0.50, 0.50{]}}}} & \textbf{\makecell{0.50 \\ {\footnotesize {[}0.50, 0.50{]}}}} & \textbf{\makecell{0.50 \\ {\footnotesize {[}0.50, 0.50{]}}}} & \textbf{\makecell{0.50 \\ {\footnotesize {[}0.50, 0.50{]}}}} \\
\makecell[r]{\textbf{AQD-Score}} & \textbf{\makecell{3 \\ {\footnotesize {[}3, 3{]}}}} & \textbf{\makecell{3 \\ {\footnotesize {[}3, 3{]}}}} & \textbf{\makecell{3 \\ {\footnotesize {[}3, 3{]}}}} & \textbf{\makecell{3 \\ {\footnotesize {[}3, 3{]}}}} & \makecell{2 \\ {\footnotesize {[}2, 2{]}}} & \makecell{2 \\ {\footnotesize {[}2, 2{]}}} & \makecell{2 \\ {\footnotesize {[}2, 2{]}}} & \makecell{2 \\ {\footnotesize {[}2, 2{]}}} \\
\bottomrule
\end{NiceTabular}
\end{table*}

\begin{table*}[]
\caption{Cat-and-mouse comparison using \textbf{Ranking} to select the competing elites for each variant.}
\label{tab:ranking_cat_and_mouse_comparison}
\vspace{-0.4cm}
\footnotesize
\begin{NiceTabular}{@{}>{\raggedleft\arraybackslash}m{2.2cm}cccccccc@{}}
\CodeBefore
  \rowcolor{gray!15}{4}
  \rowcolor{gray!15}{6}
  \rowcolor{gray!15}{8}
\Body
\toprule
 & \multicolumn{2}{c}{\textbf{Ranking}} & \multicolumn{2}{c}{\textbf{Pareto}} & \multicolumn{2}{c}{\textbf{Behavior}} & \multicolumn{2}{c}{\textbf{Random}} \\
 & Cat & Mouse & Cat & Mouse & Cat & Mouse & Cat & Mouse \\
\midrule
\makecell[r]{\textbf{Win rate}} & \textbf{\makecell{56.3\% \\ {\footnotesize {[}50.8\%, 60.1\%{]}}}} & \textbf{\makecell{86.9\% \\ {\footnotesize {[}86.1\%, 87.6\%{]}}}} & \textbf{\makecell{55.8\% \\ {\footnotesize {[}50.6\%, 60.4\%{]}}}} & \textbf{\makecell{88.2\% \\ {\footnotesize {[}86.5\%, 88.9\%{]}}}} & \makecell{41.5\% \\ {\footnotesize {[}36.1\%, 46.4\%{]}}} & \makecell{84.4\% \\ {\footnotesize {[}83.8\%, 84.8\%{]}}} & \makecell{38.5\% \\ {\footnotesize {[}34.8\%, 44.4\%{]}}} & \makecell{84.3\% \\ {\footnotesize {[}83.9\%, 85.0\%{]}}} \\
\makecell[r]{\textbf{ELO Score}} & \textbf{\makecell{99.2\% \\ {\footnotesize {[}98.1\%, 99.7\%{]}}}} & \textbf{\makecell{98.9\% \\ {\footnotesize {[}97.8\%, 99.4\%{]}}}} & \textbf{\makecell{99.1\% \\ {\footnotesize {[}98.0\%, 99.7\%{]}}}} & \textbf{\makecell{99.5\% \\ {\footnotesize {[}98.6\%, 99.7\%{]}}}} & \makecell{93.1\% \\ {\footnotesize {[}87.3\%, 96.0\%{]}}} & \makecell{90.9\% \\ {\footnotesize {[}87.6\%, 93.0\%{]}}} & \makecell{90.2\% \\ {\footnotesize {[}85.5\%, 94.9\%{]}}} & \makecell{90.4\% \\ {\footnotesize {[}87.9\%, 94.0\%{]}}} \\
\makecell[r]{\textbf{Robustness}} & \textbf{\makecell{0.40 \\ {\footnotesize {[}0.39, 0.41{]}}}} & \makecell{0.17 \\ {\footnotesize {[}0.17, 0.18{]}}} & \textbf{\makecell{0.40 \\ {\footnotesize {[}0.39, 0.42{]}}}} & \makecell{0.18 \\ {\footnotesize {[}0.17, 0.18{]}}} & \makecell{0.35 \\ {\footnotesize {[}0.33, 0.36{]}}} & \textbf{\makecell{0.18 \\ {\footnotesize {[}0.18, 0.18{]}}}} & \makecell{0.35 \\ {\footnotesize {[}0.33, 0.36{]}}} & \textbf{\makecell{0.18 \\ {\footnotesize {[}0.18, 0.18{]}}}} \\
\makecell[r]{\textbf{Coverage}} & \textbf{\makecell{61.0\% \\ {\footnotesize {[}58.0\%, 64.5\%{]}}}} & \textbf{\makecell{62.0\% \\ {\footnotesize {[}58.0\%, 64.0\%{]}}}} & \textbf{\makecell{59.0\% \\ {\footnotesize {[}54.0\%, 62.0\%{]}}}} & \makecell{56.0\% \\ {\footnotesize {[}54.0\%, 60.0\%{]}}} & \makecell{48.0\% \\ {\footnotesize {[}46.0\%, 54.0\%{]}}} & \makecell{48.0\% \\ {\footnotesize {[}39.5\%, 50.0\%{]}}} & \makecell{57.0\% \\ {\footnotesize {[}51.5\%, 58.0\%{]}}} & \makecell{54.0\% \\ {\footnotesize {[}52.0\%, 58.5\%{]}}} \\
\makecell[r]{\textbf{Expertise}} & \textbf{\makecell{0.81 \\ {\footnotesize {[}0.75, 0.82{]}}}} & \textbf{\makecell{0.55 \\ {\footnotesize {[}0.55, 0.56{]}}}} & \makecell{0.71 \\ {\footnotesize {[}0.61, 0.80{]}}} & \makecell{0.55 \\ {\footnotesize {[}0.55, 0.55{]}}} & \makecell{0.47 \\ {\footnotesize {[}0.45, 0.48{]}}} & \makecell{0.53 \\ {\footnotesize {[}0.52, 0.54{]}}} & \makecell{0.48 \\ {\footnotesize {[}0.47, 0.49{]}}} & \makecell{0.54 \\ {\footnotesize {[}0.53, 0.54{]}}} \\
\makecell[r]{\textbf{AQD-Score}} & \textbf{\makecell{2 \\ {\footnotesize {[}2, 2{]}}}} & \textbf{\makecell{3 \\ {\footnotesize {[}3, 3{]}}}} & \makecell{2 \\ {\footnotesize {[}2, 2{]}}} & \textbf{\makecell{3 \\ {\footnotesize {[}3, 3{]}}}} & \makecell{1 \\ {\footnotesize {[}1, 1{]}}} & \makecell{2 \\ {\footnotesize {[}2, 2{]}}} & \makecell{1 \\ {\footnotesize {[}1, 1{]}}} & \makecell{2 \\ {\footnotesize {[}2, 2{]}}} \\
\bottomrule
\end{NiceTabular}
\end{table*}

\begin{table*}[]
\caption{Pursuers-and-evaders comparison using \textbf{Ranking} to select the competing elites for each variant.}
\label{tab:ranking_pursuers_evaders_comparison}
\vspace{-0.4cm}
\footnotesize
\begin{NiceTabular}{@{}>{\raggedleft\arraybackslash}m{2.2cm}cccccccc@{}}
\CodeBefore
  \rowcolor{gray!15}{4}
  \rowcolor{gray!15}{6}
  \rowcolor{gray!15}{8}
\Body
\toprule
 & \multicolumn{2}{c}{\textbf{Ranking}} & \multicolumn{2}{c}{\textbf{Pareto}} & \multicolumn{2}{c}{\textbf{Behavior}} & \multicolumn{2}{c}{\textbf{Random}} \\
 & Pursuers & Evaders & Pursuers & Evaders & Pursuers & Evaders & Pursuers & Evaders \\
\midrule
\makecell[r]{\textbf{Win rate}} & \textbf{\makecell{48.3\% \\ {\footnotesize {[}46.2\%, 51.0\%{]}}}} & \textbf{\makecell{91.8\% \\ {\footnotesize {[}90.9\%, 92.4\%{]}}}} & \makecell{46.1\% \\ {\footnotesize {[}43.0\%, 48.0\%{]}}} & \makecell{90.6\% \\ {\footnotesize {[}90.2\%, 91.2\%{]}}} & \makecell{38.4\% \\ {\footnotesize {[}36.9\%, 42.5\%{]}}} & \makecell{90.6\% \\ {\footnotesize {[}88.9\%, 91.0\%{]}}} & \makecell{41.9\% \\ {\footnotesize {[}38.5\%, 43.6\%{]}}} & \makecell{89.0\% \\ {\footnotesize {[}88.4\%, 90.1\%{]}}} \\
\makecell[r]{\textbf{ELO Score}} & \textbf{\makecell{99.4\% \\ {\footnotesize {[}98.6\%, 99.9\%{]}}}} & \textbf{\makecell{99.4\% \\ {\footnotesize {[}98.7\%, 99.8\%{]}}}} & \makecell{98.6\% \\ {\footnotesize {[}96.5\%, 99.2\%{]}}} & \makecell{98.2\% \\ {\footnotesize {[}97.3\%, 99.0\%{]}}} & \makecell{89.1\% \\ {\footnotesize {[}84.5\%, 95.7\%{]}}} & \makecell{98.1\% \\ {\footnotesize {[}94.8\%, 98.8\%{]}}} & \makecell{95.1\% \\ {\footnotesize {[}89.1\%, 97.1\%{]}}} & \makecell{95.0\% \\ {\footnotesize {[}93.6\%, 97.1\%{]}}} \\
\makecell[r]{\textbf{Robustness}} & \textbf{\makecell{0.13 \\ {\footnotesize {[}0.12, 0.15{]}}}} & \makecell{0.12 \\ {\footnotesize {[}0.12, 0.12{]}}} & \textbf{\makecell{0.13 \\ {\footnotesize {[}0.12, 0.14{]}}}} & \makecell{0.12 \\ {\footnotesize {[}0.12, 0.13{]}}} & \textbf{\makecell{0.13 \\ {\footnotesize {[}0.12, 0.14{]}}}} & \textbf{\makecell{0.13 \\ {\footnotesize {[}0.12, 0.13{]}}}} & \textbf{\makecell{0.13 \\ {\footnotesize {[}0.12, 0.14{]}}}} & \makecell{0.12 \\ {\footnotesize {[}0.12, 0.12{]}}} \\
\makecell[r]{\textbf{Coverage}} & \makecell{50.0\% \\ {\footnotesize {[}47.5\%, 52.0\%{]}}} & \makecell{53.0\% \\ {\footnotesize {[}50.0\%, 56.5\%{]}}} & \makecell{45.0\% \\ {\footnotesize {[}40.0\%, 50.0\%{]}}} & \makecell{48.0\% \\ {\footnotesize {[}43.5\%, 52.5\%{]}}} & \textbf{\makecell{53.0\% \\ {\footnotesize {[}50.0\%, 58.0\%{]}}}} & \makecell{54.0\% \\ {\footnotesize {[}52.0\%, 56.0\%{]}}} & \textbf{\makecell{56.0\% \\ {\footnotesize {[}53.5\%, 60.0\%{]}}}} & \textbf{\makecell{58.0\% \\ {\footnotesize {[}55.5\%, 62.0\%{]}}}} \\
\makecell[r]{\textbf{Expertise}} & \textbf{\makecell{0.58 \\ {\footnotesize {[}0.49, 0.75{]}}}} & \textbf{\makecell{0.76 \\ {\footnotesize {[}0.60, 0.77{]}}}} & \makecell{0.49 \\ {\footnotesize {[}0.48, 0.49{]}}} & \textbf{\makecell{0.76 \\ {\footnotesize {[}0.62, 0.77{]}}}} & \makecell{0.49 \\ {\footnotesize {[}0.48, 0.49{]}}} & \makecell{0.56 \\ {\footnotesize {[}0.55, 0.57{]}}} & \makecell{0.49 \\ {\footnotesize {[}0.48, 0.49{]}}} & \makecell{0.59 \\ {\footnotesize {[}0.56, 0.76{]}}} \\
\makecell[r]{\textbf{AQD-Score}} & \textbf{\makecell{1.5 \\ {\footnotesize {[}1, 2{]}}}} & \textbf{\makecell{3 \\ {\footnotesize {[}3, 3{]}}}} & \makecell{1 \\ {\footnotesize {[}1, 1{]}}} & \textbf{\makecell{3 \\ {\footnotesize {[}3, 3{]}}}} & \makecell{1 \\ {\footnotesize {[}1, 1{]}}} & \textbf{\makecell{3 \\ {\footnotesize {[}3, 3{]}}}} & \makecell{1 \\ {\footnotesize {[}1, 1{]}}} & \textbf{\makecell{3 \\ {\footnotesize {[}3, 3{]}}}} \\
\bottomrule
\end{NiceTabular}
\end{table*}

\FloatBarrier

\section{Dominated Novelty Search}
\label{apx:DNS}

As suggested by a reviewer, after submission, we compared against an additional variant that uses Dominated Novelty Search (DNS)~\citep{bahlous2025dominated}. DNS is a recent QD variant that has shown success in better covering arbitrary-shaped behavior spaces. Compared to the default \textbf{Ranking (GA)}, which uses a Growing Archive, \textbf{Ranking (DNS)} uses DNS in each task archive of the multi-task MAP-Elites with $k$ parameter set to $\num{5}$ as suggested by the authors.

\begin{figure*}[ht]
    \centering
    \includegraphics[width=0.8\linewidth]{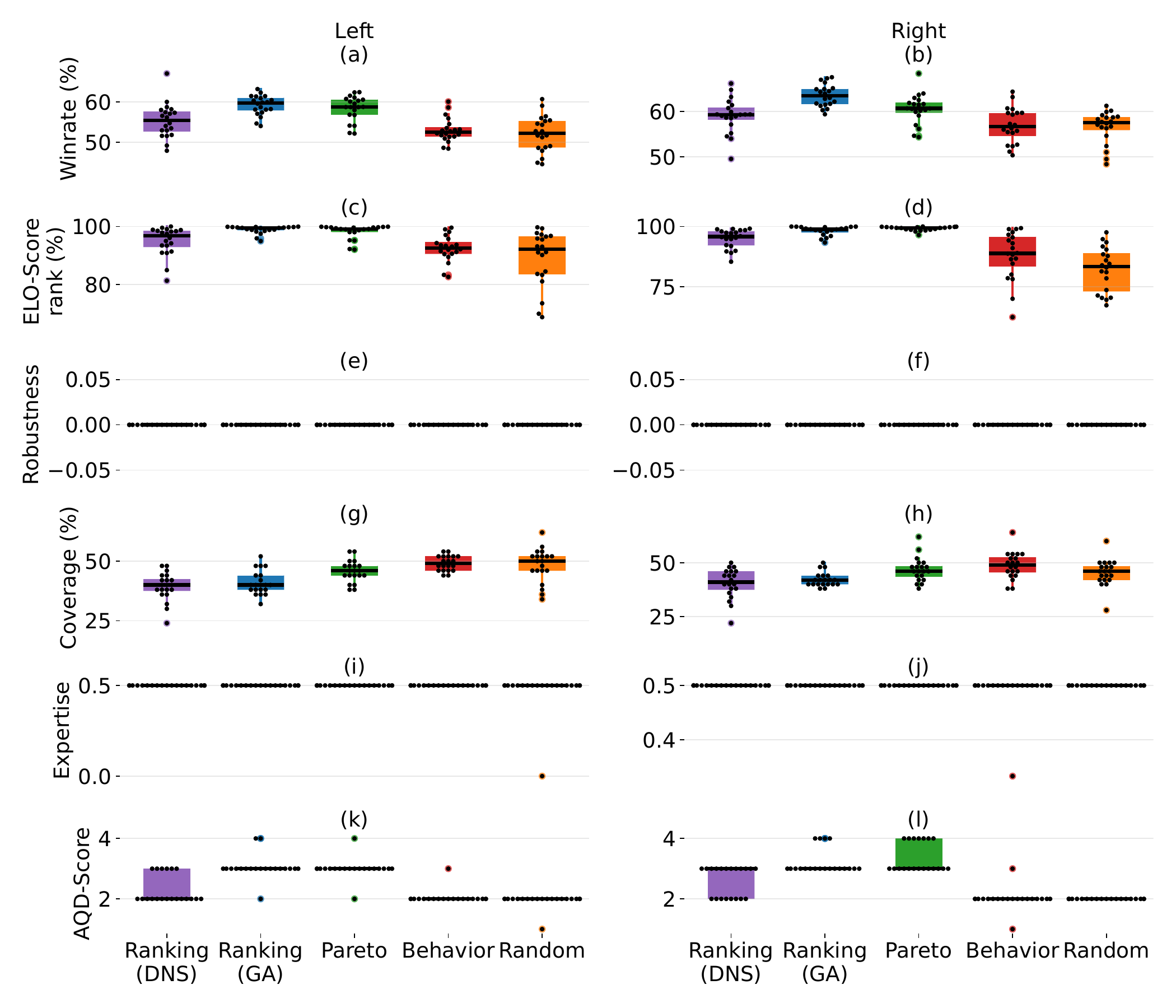}
    \caption{Comparison with \textbf{Ranking (DNS)} in Pong.}
    \label{fig:DNS_pong_comparison}
\end{figure*}

\begin{figure*}[ht]
    \centering
    \includegraphics[width=0.8\linewidth]{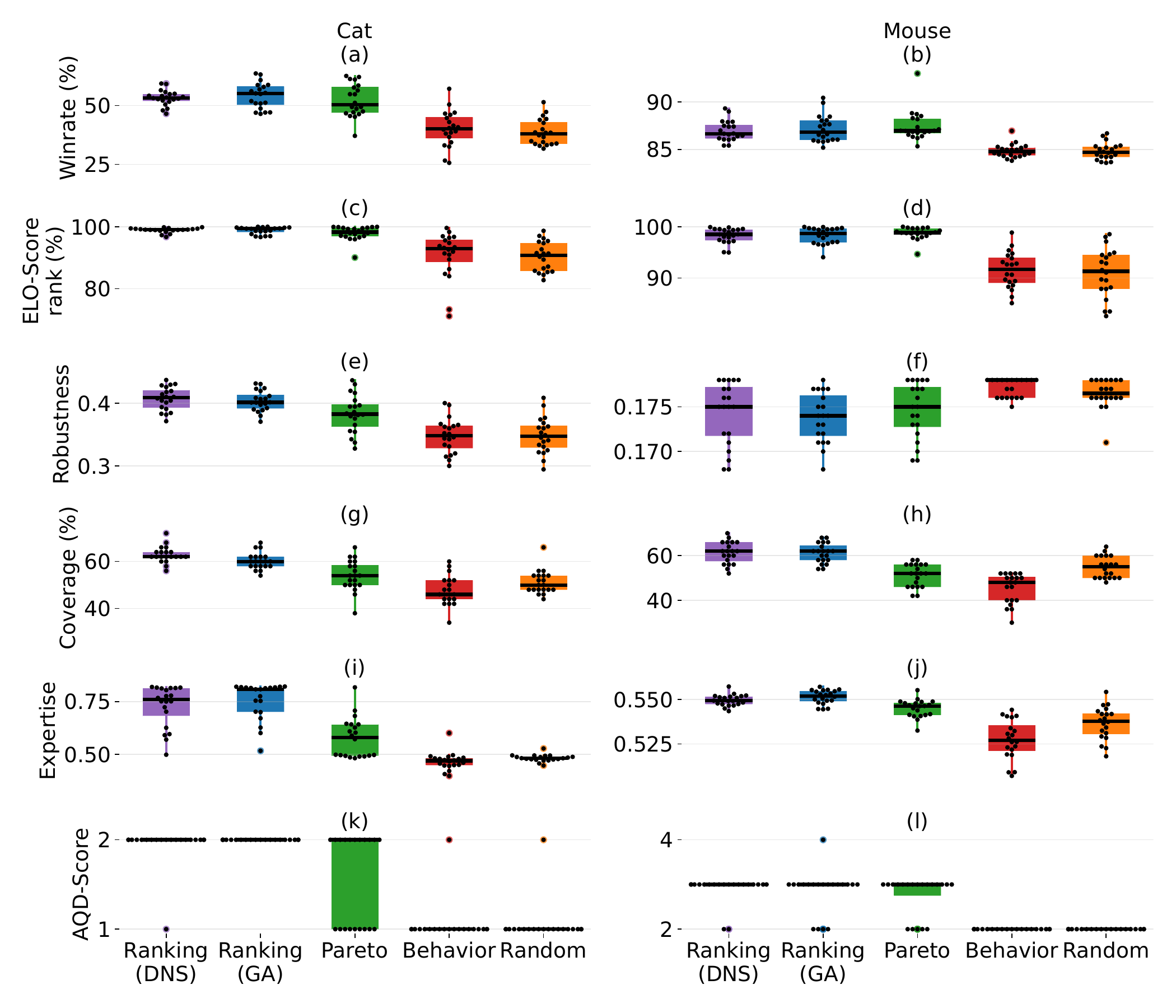}
    \caption{Comparison with \textbf{Ranking (DNS)} in Cat-and-mouse.}
    \label{fig:DNS_cat_and_mouse_comparison}
\end{figure*}

\begin{figure*}[ht]
    \centering
    \includegraphics[width=0.8\linewidth]{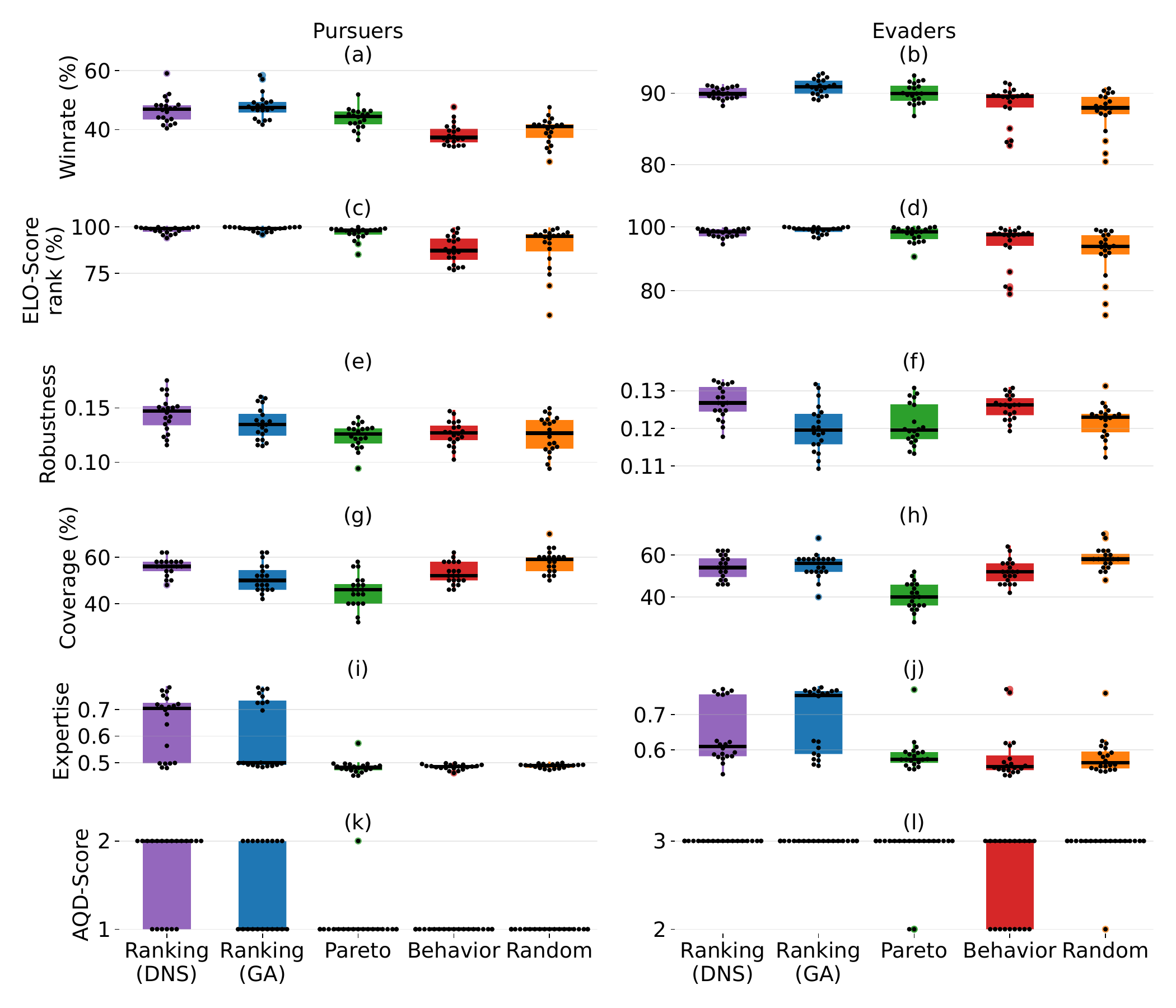}
    \caption{Pursuers-and-evaders comparison with \textbf{Ranking (DNS)}.}
    \label{fig:DNS_pursuers_and_evaders_comparison}
\end{figure*}

Fig.~\ref{fig:DNS_pong_comparison} and Tab.~\ref{tab:DNS_pong} show the result in Pong.
Fig.~\ref{fig:DNS_cat_and_mouse_comparison} and Tab.~\ref{tab:DNS_cat_and_mouse} show the result in Cat-and-mouse.
Fig.~\ref{fig:DNS_pursuers_and_evaders_comparison} and Tab.~\ref{tab:DNS_pursuers_and_evaders} show the result in Pursuers-and-evaders.

Overall, \textbf{Ranking (GA)} is significantly better in quality (Win rate and ELO Score) in Pong and the Evaders' side of Pursuers-and-evaders than \textbf{Ranking (DNS)}. \textbf{Ranking (DNS)} is significantly better in diversity: Coverage on the Cat's side in Cat-and-mouse and the Pursuers' side in Pursuers-and-evaders; and Robustness in the Evaders' side of Pursuers-and-evaders. 

This shows DNS's better handling of the large VEM behavior space for diversity at the expense of reduced quality. One can then choose the method depending on their desired goal for the illumination. 

\begin{table*}[]
\caption{Combined results in Pong with \textbf{Ranking (DNS)}}
\label{tab:DNS_pong}
\vspace{-0.4cm}
\small
\begin{NiceTabular}{@{}lcccccccccc@{}}
\CodeBefore
  \rowcolor{gray!15}{4}
  \rowcolor{gray!15}{6}
  \rowcolor{gray!15}{8}
\Body
\toprule
 & \multicolumn{2}{c}{\textbf{Ranking
(DNS)}} & \multicolumn{2}{c}{\textbf{Ranking
(GA)}} & \multicolumn{2}{c}{\textbf{Pareto}} & \multicolumn{2}{c}{\textbf{Behavior}} & \multicolumn{2}{c}{\textbf{Random}} \\
 & Left & Right & Left & Right & Left & Right & Left & Right & Left & Right \\
\midrule
\makecell[r]{\textbf{Win rate}} & \makecell{55.4\% \\ {\scriptsize {[}52.6\%, 57.6\%{]}}} & \makecell{59.3\% \\ {\scriptsize {[}58.2\%, 60.9\%{]}}} & \textbf{\makecell{59.7\% \\ {\scriptsize {[}57.9\%, 61.0\%{]}}}} & \textbf{\makecell{63.5\% \\ {\scriptsize {[}61.7\%, 65.0\%{]}}}} & \textbf{\makecell{58.7\% \\ {\scriptsize {[}56.8\%, 60.6\%{]}}}} & \makecell{60.7\% \\ {\scriptsize {[}59.7\%, 62.1\%{]}}} & \makecell{52.5\% \\ {\scriptsize {[}51.4\%, 53.8\%{]}}} & \makecell{56.7\% \\ {\scriptsize {[}54.7\%, 59.7\%{]}}} & \makecell{52.2\% \\ {\scriptsize {[}48.7\%, 55.3\%{]}}} & \makecell{57.5\% \\ {\scriptsize {[}55.9\%, 58.8\%{]}}} \\
\makecell[r]{\textbf{ELO Score}} & \makecell{96.8\% \\ {\scriptsize {[}92.9\%, 98.6\%{]}}} & \makecell{95.8\% \\ {\scriptsize {[}92.2\%, 98.1\%{]}}} & \textbf{\makecell{99.5\% \\ {\scriptsize {[}98.7\%, 99.8\%{]}}}} & \textbf{\makecell{98.8\% \\ {\scriptsize {[}97.5\%, 99.8\%{]}}}} & \textbf{\makecell{99.2\% \\ {\scriptsize {[}98.1\%, 99.7\%{]}}}} & \textbf{\makecell{99.4\% \\ {\scriptsize {[}98.7\%, 99.7\%{]}}}} & \makecell{92.6\% \\ {\scriptsize {[}90.6\%, 94.7\%{]}}} & \makecell{88.8\% \\ {\scriptsize {[}83.4\%, 95.8\%{]}}} & \makecell{92.1\% \\ {\scriptsize {[}83.6\%, 96.6\%{]}}} & \makecell{83.4\% \\ {\scriptsize {[}73.1\%, 89.0\%{]}}} \\
\makecell[r]{\textbf{Robustness}} & \textbf{\makecell{0.00 \\ {\scriptsize {[}0.00, 0.00{]}}}} & \textbf{\makecell{0.00 \\ {\scriptsize {[}0.00, 0.00{]}}}} & \textbf{\makecell{0.00 \\ {\scriptsize {[}0.00, 0.00{]}}}} & \textbf{\makecell{0.00 \\ {\scriptsize {[}0.00, 0.00{]}}}} & \textbf{\makecell{0.00 \\ {\scriptsize {[}0.00, 0.00{]}}}} & \textbf{\makecell{0.00 \\ {\scriptsize {[}0.00, 0.00{]}}}} & \textbf{\makecell{0.00 \\ {\scriptsize {[}0.00, 0.00{]}}}} & \textbf{\makecell{0.00 \\ {\scriptsize {[}0.00, 0.00{]}}}} & \textbf{\makecell{0.00 \\ {\scriptsize {[}0.00, 0.00{]}}}} & \textbf{\makecell{0.00 \\ {\scriptsize {[}0.00, 0.00{]}}}} \\
\makecell[r]{\textbf{Coverage}} & \makecell{40.0\% \\ {\scriptsize {[}37.5\%, 42.5\%{]}}} & \makecell{41.0\% \\ {\scriptsize {[}37.5\%, 46.0\%{]}}} & \makecell{40.0\% \\ {\scriptsize {[}38.0\%, 44.0\%{]}}} & \makecell{42.0\% \\ {\scriptsize {[}40.0\%, 44.0\%{]}}} & \textbf{\makecell{46.0\% \\ {\scriptsize {[}44.0\%, 48.0\%{]}}}} & \textbf{\makecell{46.0\% \\ {\scriptsize {[}43.5\%, 48.5\%{]}}}} & \textbf{\makecell{49.0\% \\ {\scriptsize {[}46.0\%, 52.0\%{]}}}} & \textbf{\makecell{49.0\% \\ {\scriptsize {[}45.5\%, 52.5\%{]}}}} & \textbf{\makecell{50.0\% \\ {\scriptsize {[}46.0\%, 52.0\%{]}}}} & \textbf{\makecell{46.0\% \\ {\scriptsize {[}42.0\%, 48.5\%{]}}}} \\
\makecell[r]{\textbf{Expertise}} & \textbf{\makecell{0.50 \\ {\scriptsize {[}0.50, 0.50{]}}}} & \textbf{\makecell{0.50 \\ {\scriptsize {[}0.50, 0.50{]}}}} & \textbf{\makecell{0.50 \\ {\scriptsize {[}0.50, 0.50{]}}}} & \textbf{\makecell{0.50 \\ {\scriptsize {[}0.50, 0.50{]}}}} & \textbf{\makecell{0.50 \\ {\scriptsize {[}0.50, 0.50{]}}}} & \textbf{\makecell{0.50 \\ {\scriptsize {[}0.50, 0.50{]}}}} & \textbf{\makecell{0.50 \\ {\scriptsize {[}0.50, 0.50{]}}}} & \textbf{\makecell{0.50 \\ {\scriptsize {[}0.50, 0.50{]}}}} & \textbf{\makecell{0.50 \\ {\scriptsize {[}0.50, 0.50{]}}}} & \textbf{\makecell{0.50 \\ {\scriptsize {[}0.50, 0.50{]}}}} \\
\makecell[r]{\textbf{AQD-Score}} & \makecell{2 \\ {\scriptsize {[}2, 3{]}}} & \textbf{\makecell{3 \\ {\scriptsize {[}2, 3{]}}}} & \textbf{\makecell{3 \\ {\scriptsize {[}3, 3{]}}}} & \makecell{3 \\ {\scriptsize {[}3, 3{]}}} & \textbf{\makecell{3 \\ {\scriptsize {[}3, 3{]}}}} & \makecell{3 \\ {\scriptsize {[}3, 4{]}}} & \makecell{2 \\ {\scriptsize {[}2, 2{]}}} & \makecell{2 \\ {\scriptsize {[}2, 2{]}}} & \makecell{2 \\ {\scriptsize {[}2, 2{]}}} & \makecell{2 \\ {\scriptsize {[}2, 2{]}}} \\
\bottomrule
\end{NiceTabular}
\end{table*}

\begin{table*}[]
\caption{Combined results in Cat-and-mouse with \textbf{Ranking (DNS)}}
\label{tab:DNS_cat_and_mouse}
\vspace{-0.4cm}
\small
\begin{NiceTabular}{@{}lcccccccccc@{}}
\CodeBefore
  \rowcolor{gray!15}{4}
  \rowcolor{gray!15}{6}
  \rowcolor{gray!15}{8}
\Body
\toprule
 & \multicolumn{2}{c}{\textbf{Ranking
(DNS)}} & \multicolumn{2}{c}{\textbf{Ranking
(GA)}} & \multicolumn{2}{c}{\textbf{Pareto}} & \multicolumn{2}{c}{\textbf{Behavior}} & \multicolumn{2}{c}{\textbf{Random}} \\
 & Cat & Mouse & Cat & Mouse & Cat & Mouse & Cat & Mouse & Cat & Mouse \\
\midrule
\makecell[r]{\textbf{Win rate}} & \textbf{\makecell{53.3\% \\ {\scriptsize {[}52.0\%, 54.9\%{]}}}} & \textbf{\makecell{86.7\% \\ {\scriptsize {[}86.1\%, 87.6\%{]}}}} & \textbf{\makecell{55.1\% \\ {\scriptsize {[}50.4\%, 58.2\%{]}}}} & \textbf{\makecell{86.8\% \\ {\scriptsize {[}86.0\%, 88.0\%{]}}}} & \textbf{\makecell{50.3\% \\ {\scriptsize {[}47.0\%, 57.9\%{]}}}} & \textbf{\makecell{87.0\% \\ {\scriptsize {[}86.7\%, 88.2\%{]}}}} & \makecell{40.2\% \\ {\scriptsize {[}36.1\%, 45.1\%{]}}} & \makecell{84.8\% \\ {\scriptsize {[}84.4\%, 85.2\%{]}}} & \makecell{37.9\% \\ {\scriptsize {[}33.8\%, 43.0\%{]}}} & \makecell{84.7\% \\ {\scriptsize {[}84.2\%, 85.3\%{]}}} \\
\makecell[r]{\textbf{ELO Score}} & \textbf{\makecell{99.0\% \\ {\scriptsize {[}98.8\%, 99.3\%{]}}}} & \textbf{\makecell{98.5\% \\ {\scriptsize {[}97.4\%, 99.5\%{]}}}} & \textbf{\makecell{99.4\% \\ {\scriptsize {[}98.3\%, 99.7\%{]}}}} & \textbf{\makecell{98.7\% \\ {\scriptsize {[}97.0\%, 99.6\%{]}}}} & \textbf{\makecell{98.3\% \\ {\scriptsize {[}96.9\%, 99.7\%{]}}}} & \textbf{\makecell{98.8\% \\ {\scriptsize {[}98.5\%, 99.7\%{]}}}} & \makecell{92.9\% \\ {\scriptsize {[}88.7\%, 95.8\%{]}}} & \makecell{91.7\% \\ {\scriptsize {[}89.1\%, 94.0\%{]}}} & \makecell{90.8\% \\ {\scriptsize {[}85.7\%, 94.8\%{]}}} & \makecell{91.3\% \\ {\scriptsize {[}87.9\%, 94.5\%{]}}} \\
\makecell[r]{\textbf{Robustness}} & \textbf{\makecell{0.41 \\ {\scriptsize {[}0.39, 0.42{]}}}} & \makecell{0.18 \\ {\scriptsize {[}0.17, 0.18{]}}} & \textbf{\makecell{0.40 \\ {\scriptsize {[}0.39, 0.41{]}}}} & \makecell{0.17 \\ {\scriptsize {[}0.17, 0.18{]}}} & \makecell{0.38 \\ {\scriptsize {[}0.36, 0.40{]}}} & \makecell{0.18 \\ {\scriptsize {[}0.17, 0.18{]}}} & \makecell{0.35 \\ {\scriptsize {[}0.33, 0.36{]}}} & \textbf{\makecell{0.18 \\ {\scriptsize {[}0.18, 0.18{]}}}} & \makecell{0.35 \\ {\scriptsize {[}0.33, 0.36{]}}} & \textbf{\makecell{0.18 \\ {\scriptsize {[}0.18, 0.18{]}}}} \\
\makecell[r]{\textbf{Coverage}} & \textbf{\makecell{62.0\% \\ {\scriptsize {[}62.0\%, 64.0\%{]}}}} & \textbf{\makecell{62.0\% \\ {\scriptsize {[}57.5\%, 66.0\%{]}}}} & \makecell{60.0\% \\ {\scriptsize {[}58.0\%, 62.0\%{]}}} & \textbf{\makecell{62.0\% \\ {\scriptsize {[}58.0\%, 64.5\%{]}}}} & \makecell{54.0\% \\ {\scriptsize {[}50.0\%, 58.5\%{]}}} & \makecell{52.0\% \\ {\scriptsize {[}46.0\%, 56.0\%{]}}} & \makecell{46.0\% \\ {\scriptsize {[}44.0\%, 52.0\%{]}}} & \makecell{48.0\% \\ {\scriptsize {[}40.0\%, 50.5\%{]}}} & \makecell{50.0\% \\ {\scriptsize {[}48.0\%, 54.0\%{]}}} & \makecell{55.0\% \\ {\scriptsize {[}50.0\%, 60.0\%{]}}} \\
\makecell[r]{\textbf{Expertise}} & \textbf{\makecell{0.76 \\ {\scriptsize {[}0.68, 0.81{]}}}} & \textbf{\makecell{0.55 \\ {\scriptsize {[}0.55, 0.55{]}}}} & \textbf{\makecell{0.81 \\ {\scriptsize {[}0.70, 0.82{]}}}} & \textbf{\makecell{0.55 \\ {\scriptsize {[}0.55, 0.55{]}}}} & \makecell{0.58 \\ {\scriptsize {[}0.49, 0.64{]}}} & \makecell{0.55 \\ {\scriptsize {[}0.54, 0.55{]}}} & \makecell{0.47 \\ {\scriptsize {[}0.45, 0.48{]}}} & \makecell{0.53 \\ {\scriptsize {[}0.52, 0.54{]}}} & \makecell{0.48 \\ {\scriptsize {[}0.48, 0.49{]}}} & \makecell{0.54 \\ {\scriptsize {[}0.53, 0.54{]}}} \\
\makecell[r]{\textbf{AQD-Score}} & \textbf{\makecell{2 \\ {\scriptsize {[}2, 2{]}}}} & \textbf{\makecell{3 \\ {\scriptsize {[}3, 3{]}}}} & \textbf{\makecell{2 \\ {\scriptsize {[}2, 2{]}}}} & \textbf{\makecell{3 \\ {\scriptsize {[}3, 3{]}}}} & \makecell{2 \\ {\scriptsize {[}1, 2{]}}} & \textbf{\makecell{3 \\ {\scriptsize {[}2.8, 3{]}}}} & \makecell{1 \\ {\scriptsize {[}1, 1{]}}} & \makecell{2 \\ {\scriptsize {[}2, 2{]}}} & \makecell{1 \\ {\scriptsize {[}1, 1{]}}} & \makecell{2 \\ {\scriptsize {[}2, 2{]}}} \\
\bottomrule
\end{NiceTabular}
\end{table*}

\begin{table*}[!hbp]
\caption{Combined results for Pursuers-and-evaders with \textbf{Ranking (DNS)}.}
\label{tab:DNS_pursuers_and_evaders}
\vspace{-0.4cm}
\small
\begin{NiceTabular}{@{}lcccccccccc@{}}
\CodeBefore
  \rowcolor{gray!15}{4}
  \rowcolor{gray!15}{6}
  \rowcolor{gray!15}{8}
\Body
\toprule
 & \multicolumn{2}{c}{\textbf{Ranking
(DNS)}} & \multicolumn{2}{c}{\textbf{Ranking
(GA)}} & \multicolumn{2}{c}{\textbf{Pareto}} & \multicolumn{2}{c}{\textbf{Behavior}} & \multicolumn{2}{c}{\textbf{Random}} \\
 & Pursuers & Evaders & Pursuers & Evaders & Pursuers & Evaders & Pursuers & Evaders & Pursuers & Evaders \\
\midrule
\makecell[r]{\textbf{Win rate}} & \textbf{\makecell{46.9\% \\ {\scriptsize {[}43.4\%, 48.3\%{]}}}} & \makecell{89.9\% \\ {\scriptsize {[}89.3\%, 90.7\%{]}}} & \textbf{\makecell{47.4\% \\ {\scriptsize {[}45.8\%, 49.3\%{]}}}} & \textbf{\makecell{90.9\% \\ {\scriptsize {[}89.9\%, 91.7\%{]}}}} & \makecell{44.5\% \\ {\scriptsize {[}41.8\%, 46.1\%{]}}} & \textbf{\makecell{89.9\% \\ {\scriptsize {[}88.9\%, 91.0\%{]}}}} & \makecell{37.3\% \\ {\scriptsize {[}35.6\%, 40.2\%{]}}} & \makecell{89.5\% \\ {\scriptsize {[}88.0\%, 89.8\%{]}}} & \makecell{41.0\% \\ {\scriptsize {[}37.2\%, 41.9\%{]}}} & \makecell{88.0\% \\ {\scriptsize {[}87.1\%, 89.5\%{]}}} \\
\makecell[r]{\textbf{ELO Score}} & \textbf{\makecell{99.1\% \\ {\scriptsize {[}97.5\%, 99.5\%{]}}}} & \makecell{98.4\% \\ {\scriptsize {[}97.1\%, 99.3\%{]}}} & \textbf{\makecell{99.2\% \\ {\scriptsize {[}98.6\%, 99.7\%{]}}}} & \textbf{\makecell{99.4\% \\ {\scriptsize {[}98.4\%, 99.8\%{]}}}} & \makecell{98.1\% \\ {\scriptsize {[}95.9\%, 98.8\%{]}}} & \textbf{\makecell{98.4\% \\ {\scriptsize {[}96.2\%, 99.5\%{]}}}} & \makecell{87.3\% \\ {\scriptsize {[}82.4\%, 93.7\%{]}}} & \makecell{97.6\% \\ {\scriptsize {[}94.1\%, 98.2\%{]}}} & \makecell{94.9\% \\ {\scriptsize {[}86.9\%, 96.1\%{]}}} & \makecell{93.8\% \\ {\scriptsize {[}91.3\%, 97.4\%{]}}} \\
\makecell[r]{\textbf{Robustness}} & \textbf{\makecell{0.15 \\ {\scriptsize {[}0.13, 0.15{]}}}} & \textbf{\makecell{0.13 \\ {\scriptsize {[}0.12, 0.13{]}}}} & \textbf{\makecell{0.13 \\ {\scriptsize {[}0.12, 0.14{]}}}} & \makecell{0.12 \\ {\scriptsize {[}0.12, 0.12{]}}} & \makecell{0.13 \\ {\scriptsize {[}0.12, 0.13{]}}} & \makecell{0.12 \\ {\scriptsize {[}0.12, 0.13{]}}} & \makecell{0.13 \\ {\scriptsize {[}0.12, 0.13{]}}} & \textbf{\makecell{0.13 \\ {\scriptsize {[}0.12, 0.13{]}}}} & \makecell{0.13 \\ {\scriptsize {[}0.11, 0.14{]}}} & \makecell{0.12 \\ {\scriptsize {[}0.12, 0.12{]}}} \\
\makecell[r]{\textbf{Coverage}} & \textbf{\makecell{56.0\% \\ {\scriptsize {[}54.0\%, 58.0\%{]}}}} & \textbf{\makecell{54.0\% \\ {\scriptsize {[}49.5\%, 58.5\%{]}}}} & \makecell{50.0\% \\ {\scriptsize {[}46.0\%, 54.5\%{]}}} & \textbf{\makecell{56.0\% \\ {\scriptsize {[}52.0\%, 58.0\%{]}}}} & \makecell{46.0\% \\ {\scriptsize {[}40.0\%, 48.5\%{]}}} & \makecell{40.0\% \\ {\scriptsize {[}36.0\%, 46.0\%{]}}} & \makecell{52.0\% \\ {\scriptsize {[}50.0\%, 58.0\%{]}}} & \makecell{52.0\% \\ {\scriptsize {[}47.5\%, 56.0\%{]}}} & \textbf{\makecell{59.0\% \\ {\scriptsize {[}54.0\%, 60.0\%{]}}}} & \textbf{\makecell{58.0\% \\ {\scriptsize {[}55.5\%, 60.5\%{]}}}} \\
\makecell[r]{\textbf{Expertise}} & \textbf{\makecell{0.70 \\ {\scriptsize {[}0.50, 0.73{]}}}} & \textbf{\makecell{0.61 \\ {\scriptsize {[}0.58, 0.76{]}}}} & \textbf{\makecell{0.50 \\ {\scriptsize {[}0.49, 0.73{]}}}} & \textbf{\makecell{0.75 \\ {\scriptsize {[}0.59, 0.77{]}}}} & \makecell{0.48 \\ {\scriptsize {[}0.47, 0.49{]}}} & \makecell{0.57 \\ {\scriptsize {[}0.56, 0.59{]}}} & \makecell{0.49 \\ {\scriptsize {[}0.48, 0.49{]}}} & \makecell{0.55 \\ {\scriptsize {[}0.54, 0.58{]}}} & \makecell{0.49 \\ {\scriptsize {[}0.48, 0.49{]}}} & \makecell{0.56 \\ {\scriptsize {[}0.55, 0.60{]}}} \\
\makecell[r]{\textbf{AQD-Score}} & \textbf{\makecell{2 \\ {\scriptsize {[}1, 2{]}}}} & \textbf{\makecell{3 \\ {\scriptsize {[}3, 3{]}}}} & \textbf{\makecell{1 \\ {\scriptsize {[}1, 2{]}}}} & \textbf{\makecell{3 \\ {\scriptsize {[}3, 3{]}}}} & \makecell{1 \\ {\scriptsize {[}1, 1{]}}} & \textbf{\makecell{3 \\ {\scriptsize {[}3, 3{]}}}} & \makecell{1 \\ {\scriptsize {[}1, 1{]}}} & \makecell{3 \\ {\scriptsize {[}2, 3{]}}} & \makecell{1 \\ {\scriptsize {[}1, 1{]}}} & \textbf{\makecell{3 \\ {\scriptsize {[}3, 3{]}}}} \\
\bottomrule
\end{NiceTabular}
\end{table*}

\end{document}